\documentclass{article}

\usepackage[preprint]{corl_2020}

\usepackage{microtype}

\usepackage{amssymb}
\usepackage{bbm}
\usepackage{amsmath}
\usepackage{pgfplots}
\usepackage{dsfont}
\usepackage{xspace}
\usepackage{multirow}

\usepackage[figuresleft]{rotating}

\usepackage{booktabs}

\usepackage{arydshln}

\usepackage{titlesec}
\titlespacing*{\section}     {0pt}{0.6ex plus 0.1ex minus .1ex}{0.4ex plus 0ex}
\titlespacing*{\subsection}   {0pt}{0.2ex plus 0.2ex minus .2ex}{0.2ex plus .1ex}
\titlespacing*{\paragraph}   {0pt}{0ex}{1em}
\newcommand{\approptoinn}[2]{\mathrel{\vcenter{
  \offinterlineskip\halign{\hfil$##$\cr
    #1\propto\cr\noalign{\kern2pt}#1\sim\cr\noalign{\kern-2pt}}}}}

\usepackage{algorithmicx} 
\usepackage[noend]{algpseudocode}
\usepackage{algorithm}
\usepackage[T1]{fontenc}

\algnewcommand\algorithmicdefinitions{\textbf{Definitions:}}
\algnewcommand\Definitions{\item[\algorithmicdefinitions]}
\renewcommand{\algorithmiccomment}[1]{{\color{gray}\raisebox{1px}{\texttt{\guillemotright}} #1}}
\algnewcommand{\LineComment}[1]{\Statex \hskip\ALG@thistlm \algorithmiccomment{#1}}
\algrenewcommand\alglinenumber[1]{\footnotesize #1:}
\algrenewcommand\algorithmicindent{1.0em}%
\makeatletter
\newcommand{\StatexIndent}[1][3]{%
  \setlength\@tempdima{\algorithmicindent}%
  \Statex\hskip\dimexpr#1\@tempdima\relax}

\makeatletter
\g@addto@macro\small{%
  \setlength\abovedisplayskip{-12pt}
  \setlength\abovedisplayshortskip{-10pt}
  \setlength\belowdisplayshortskip{-6pt}
  \setlength\belowdisplayskip{-8pt}
}
\makeatother

\newcommand{\suploss}{\mathcal{L}_{\rm SL}}

\newcommand{\domainsim}{\texttt{S}}

\newcommand{\posoob}{p^{\texttt{oob}}}
\newcommand{\dataset}{\mathcal{D}}

\newcommand{\reward}{r}

\newcommand{\stopreward}{\reward_{s}}
\newcommand{\visitreward}{\reward_{v}}

\newcommand{\actionreward}{\reward_{a}}
\newcommand{\explorereward}{\reward_{e}}

\newcommand{\stopweight}{\lambda_{s}}
\newcommand{\visitweight}{\lambda_{v}}
\newcommand{\stepweight}{\lambda_{\rm step}}
\newcommand{\actionweight}{\lambda_{a}}
\newcommand{\exploreweight}{\lambda_{e}}

\newcommand{\paramsA}{\theta}

\newcommand{\paramsB}{\phi}
\newcommand{\stageA}{f}
\newcommand{\stageAExpert}{f^*}
\newcommand{\stageAsim}{f_{\domainsim}}
\newcommand{\stageAreal}{f}
\newcommand{\stageB}{g}
\newcommand{\segmodel}{S}

\newcommand{\valueparams}{\upsilon}
\newcommand{\valuefunc}{V}

\newcommand{\eat}[1]{\ignorespaces}

\newcommand{\xxcomment}[4]{\textcolor{#1}{[${\textsc{#2#3}}$: #4]}}
\newcommand{\ya}[1]{\xxcomment{red}{Y}{A}{#1}}

\newcommand{\execlen}{T}
\newcommand{\idxtimestep}{t}

\newcommand{\trainsetsize}{N}
\newcommand{\testsetsize}{M}

\newcommand{\reals}{\mathds{R}}
\newcommand{\pdf}{pdf}

\newcommand{\state}{s}

\newcommand{\configuration}{\rho}
\newcommand{\position}{p}
\newcommand{\posseq}{\Xi}
\newcommand{\execposseq}{\hat{\Xi}}

\newcommand{\execution}{\Xi}

\newcommand{\trajvisit}{d^{\position}}
\newcommand{\stopvisit}{d^{g}}

\newcommand{\context}{c}
\newcommand{\gencontexts}{\mathcal{C}}

\newcommand{\act}[1]{{\tt #1}}
\newcommand{\action}{a}

\newcommand{\stopaction}{\act{STOP}}

\newcommand{\nlstring}[1]{{\em #1}}

\newcommand{\rail}{\textsc{SuReAL}\xspace}
\newcommand{\sureal}{\textsc{SuReAL}\xspace}
\newcommand{\fullsureal}{Supervised and Reinforcement Asynchronous Learning\xspace}

\newcommand{\oracle}{\pi^*}

\newcommand{\instruction}{u}
\newcommand{\anoninstruction}{\hat{u}}

\newcommand{\velfwd}{v}
\newcommand{\velang}{\omega}
\newcommand{\stopprob}{p^{\stopaction}}
\newcommand{\image}{I}
\newcommand{\pose}{P}
\newcommand{\nod}{\mathcal{O}}
\newcommand{\query}{Q}
\newcommand{\descr}{W}

\newcommand{\object}{o}
\newcommand{\objref}{r}

\newcommand{\objreftok}{\texttt{OBJ\_REF}}

\newcommand{\bbox}{b}
\newcommand{\objrefset}{\mathcal{R}}

\newcommand{\bboxset}{\mathcal{B}}

\newcommand{\objrefclassifier}{\textsc{ObjRef}}
\newcommand{\imageembedding}{\textsc{ImgEmb}}

\newcommand{\rpn}{\textsc{RPN}}
\newcommand{\contextemb}{\psi}

\newcommand{\worldframe}{W}

\newcommand{\segmaskworld}{\mathbf{M}^{\worldframe}}

\newcommand{\contextmapworld}{\mathbf{C}^{\worldframe}}

\newcommand{\maskworld}{\mathbf{M}^\worldframe}
\newcommand{\boundaryworld}{\mathbf{B}^\worldframe}

\newcommand{\weights}{\mathbf{W}}
\newcommand{\bias}{\mathbf{b}}

\newcommand{\lingunet}{\textsc{LingUNet}}

\newcommand{\alldata}{\textsc{ALL}}
\newcommand{\seendata}{\textsc{SEEN}}
\newcommand{\lani}{\textsc{Lani}}

\newcommand{\pvntwo}{\textsc{PVN2}}
\newcommand{\pvntwoall}{\textsc{PVN2-\alldata}}
\newcommand{\pvntwoseen}{\textsc{PVN2-\seendata}}
\newcommand{\fspvn}{\textsc{FsPVN}}

\newcommand{\fspvnBC}{\textsc{FsPVN-BC}}
\newcommand{\fspvnbignod}{\textsc{FsPVN-Big}\nod}
\newcommand{\fspvndeaf}{\textsc{FsPVN-No}\instruction}
\newcommand{\fspvnblind}{\textsc{FsPVN-No}\image}
\newcommand{\sysoracle}{\textsc{Oracle}}

\newcommand{\oraclemodel}{\textsc{Oracle}}

\newcommand{\avgmodel}{\textsc{Average}}

\newcommand{\objectdata}{\mathcal{D}_{\object}}
\newcommand{\objmask}{m}

\newcommand{\tripletloss}{\mathcal{L}_{T}}
\newcommand{\tripletmargin}{T_{M1}}
\newcommand{\tripletmarginB}{T_{M2}}

\newcommand{\featmap}{\mathbf{F}}
\newcommand{\featmaptxt}{\mathbf{G}}
\newcommand{\featmapdeconv}{\mathbf{H}}
\newcommand{\conv}{\textsc{CNN}}

\newcommand{\upscale}{\textsc{Upscale}}

\newcommand{\lingunetvecout}{\textbf{h}}
\newcommand{\avgpool}{\textsc{AvgPool}}
\newcommand{\kernel}{\mathbf{K}}

\title{Few-shot Object Grounding and Mapping \\ for Natural Language Robot Instruction Following}

\newcommand{\authorgap}{\hspace{1em}}
\author{
   Valts Blukis$^{1}$ \authorgap Ross A. Knepper\thanks{Work began while author Ross Knepper was affiliated with Cornell University.} \authorgap Yoav Artzi$^{3}$ \vspace{0.5em}\\
   $^{1,3}$Department of Computer Science and Cornell Tech, Cornell University, New York, NY, USA\\
  	\texttt{\{$^1$valts, $^3$yoav\}@cs.cornell.edu} %
}

\date{}

\begin{document}

\maketitle

\begin{abstract}
We study the problem of learning a robot policy to follow natural language instructions that can be easily extended to reason about new objects. 
We introduce a few-shot language-conditioned object grounding method trained from augmented reality data that uses exemplars to identify objects and align them to their mentions in instructions. 
We present a learned map representation that encodes object locations and their instructed use, and construct it from our few-shot grounding output. 
We integrate this mapping approach into an instruction-following policy, thereby allowing it to  reason about previously unseen objects at test-time by simply adding exemplars. 
We evaluate on the task of learning to map raw observations and instructions to continuous control of a physical quadcopter. 
Our approach significantly outperforms the prior state of the art in the presence of new objects, even when the prior approach observes all objects during training.
\end{abstract}


\section{Introduction}\label{sec:intro}

Executing natural language instructions with robotic agents requires addressing a diverse set of problems, including language understanding, perception, planning, and control. 
Most commonly, such systems are a combination of separately built modules~\cite[e.g.,][]{tellex11grounding, duvallet2013imitation, misra2014context, hemachandra2015learning, gopalan2018sequence}.  
Beyond the high engineering and integration costs of such a system, extending it, for example to reason about new object types, demands complex updates across multiple modules. 
This is also challenging in recent representation learning approaches, which learn to directly map raw observations and instructions to continuous control~\cite{blukis2019learning}. 
Learned representations entangle different aspects of the problem, making it challenging to extend model reasoning without re-training on additional data.

This paper makes two general contributions. First, we propose a few-shot method to ground natural language object mentions to their observations in the world. 
Second, we design a process to construct an object-centric learned map from groundings of object mentions within instructions. 
We show the effectiveness of this  map for instruction following by integrating it into an existing policy design to map from raw observations to continuous control. 
The policy's few-shot grounding process allows it to  reason about previously unseen objects without requiring any additional fine-tuning or training data.
The system explicitly reasons about objects and object references, while retaining the reduction in representation design and engineering that motivates learning approaches for language grounding.

\begin{figure}
    \centering
    \includegraphics[width=14cm,clip,trim=5 5 13 2]{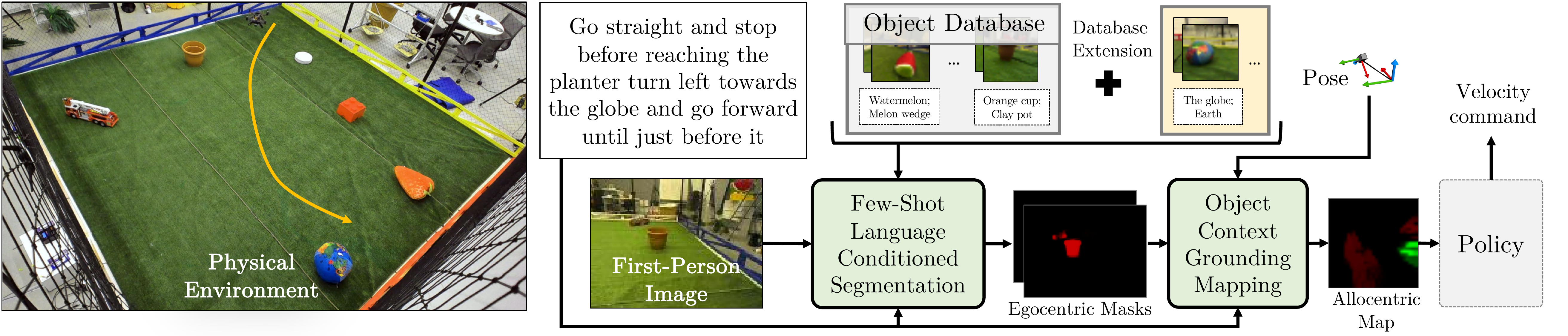}
    \vspace{-18pt}
    \caption{
    Task and approach illustration, including a third-person view of the environment (unavailable to the agent), an agent's first-person RGB observation, a natural language instruction, and an object database. The agent's reasoning can be extended by adding entries to the database. }\label{fig:intro}
    \vspace{-5pt}
\end{figure}

Figure~\ref{fig:intro} illustrates our approach.  
Rather than learning to implicitly ground instructions inside an opaque neural network model, our few-shot grounding method learns to align natural language mentions within instructions to objects in the environment using a database, which includes exemplars of object appearances and names. 
This does not require modeling specific objects or object types, but instead relies on  learning generic object properties and language similarity. 
The system's abilities are easily extended to reason about new objects by extending the database. For example, a user teaching a robot about a new object can simply take a few photos of the object and describe it. 
In contrast to existing approaches~\cite[e.g.,][]{tellex11grounding,duvallet2013imitation,hemachandra2015learning,blukis2019learning}, ours does not require additional instruction data, training, tuning, or engineering to follow instructions that mention the new object.

We train the object grounding component to recognize objects using a large augmented reality (AR) dataset of synthetic 3D objects automatically overlaid on environment images. 
This data is cheap to create, and its scale enables the model to generalize beyond the properties of any specific object. 
We train the complete policy to map instructions and inferred alignments  to continuous control in a simulation. 
Because we learn to reason about object appearance in the real environment using AR data, we can immediately deploy the model for flight in the physical environment by swapping the object grounding component  from one trained on simulator-based AR data to one trained on real-world AR data, without  any domain adaptation or training in the real world.

We evaluate on a physical quadcopter situated in an environment that contains only previously unseen objects. 
The only information available about each object is a handful of images and descriptive noun phrases.
Our approach's object-centric generalization stands in contrast to symbolic methods that typically require anticipating the full set of objects, or representation learning methods that require training data with similar objects.
Our few-shot policy outperforms the existing state of the art by 27\% absolute improvement in terms of human evaluation scores, and even outperforms a model that has seen the full set of objects during training by 10\%.
Code and videos are available at \href{https://github.com/lil-lab/drif/tree/fewshot2020}{\tt https://github.com/lil-lab/drif/tree/fewshot2020}.

\section{Related Work}\label{sec:relatedwork}

Natural language instruction following on physical robots is most commonly studied using hand-engineered symbolic  representations of world state or instruction semantics~\cite{tellex11grounding, matuszek2012joint, duvallet2013imitation, misra2014context, hemachandra2015learning, thomason2015learning, gopalan2018sequence, williams2018learning}, which require representation design and state estimation pipelines that are hard to scale to complex environments.
Recently, representation learning based on deep neural networks has been used for this task by mapping raw first-person observations and pose estimates to continuous control~\cite{blukis2019learning}.
Prior, representation learning was studied on language tasks restricted to simulated discrete~\cite{misra2017mapping, shah2018follownet, anderson2017vision, misra2018mapping, fried2018speaker, jain2019stay} and continuous~\cite{blukis2018mapping, paxton2019prospection, roh2020conditional, shridhar2020alfred} environments, or non-language robotic tasks~\cite{lenz2015deep, levine2016learning, quillen2018deep,  nair2017combining}.

Representation learning reduces the engineering effort, but results in models that are difficult to extend.  
For example, the $\pvntwo$ model by \citet{blukis2019learning} is evaluated in environments that consist of 63 different objects, all seen by the model during training. As we show in Section~\ref{sec:results}, $\pvntwo$ fails to handle new objects during test time. 
Other work has shown generalization to new indoor scenes~\cite{tan2019learning, anderson2017vision, gordon2017iqa}, but not to objects not represented in the training set.
In contrast, our representation learning approach enables deployment in environments with new objects not seen before during training, without any additional instruction data. 
We use the two-stage model decomposition, $\sureal$ training algorithm, and map projection mechanism from \citet{blukis2018following,blukis2019learning}, but completely re-design the perception, language understanding, grounding, and mapping mechanisms. 
Our system contribution is a robot representation learning system that follows natural language instructions with easily extensible reasoning capabilities. To the best of our knowledge, no existing approach provides this.

Rather than relying on new training data, we use an extensible database of visual and linguistic exemplars in a few-shot setup. 
At the core of our approach is a few-shot language-conditioned segmentation component. This mechanism is related to Prototypical Networks~\cite{snell2017prototypical}, but integrates both vision and language modalities.
Vision-only few-shot learning has been studied extensively for classification~\cite[e.g.,][]{wertheimer2019fewshot, wang2017learning, snell2017prototypical} and segmentation~\cite{gao2019ssap}.
Our language-conditioned segmentation problem is a variant of referring expression recognition~\cite{roy2019leveraging, margffoy2018dynamic, yu2018mattnet, cirik2018visual,shridhar2018interactive}. 
Our method is related to a recent alignment-based approach to referring expression resolution using an object database~\cite{roy2019leveraging}. 

\section{Technical Overview}\label{sec:tech-overview}

Our focus is reasoning about objects not seen during training. 
This overview places our work in the context of the complete system. 
We adopt the task setup, policy decomposition, and parts of the learning process from \citet{blukis2019learning}, and borrow parts of the overview for consistency.

\paragraph{Task Setup}

Our goal is to map natural language navigation instructions to continuous control of a quadcopter drone. 
The agent behavior is determined by a velocity controller setpoint $\configuration = ( \velfwd, \velang )$, where  $\velfwd\in\reals$ is a forward velocity  and  $\velang\in\reals$ is a yaw rate.
The model generates actions at fixed intervals.  
An action is either the task completion action $\stopaction$ or a setpoint update $( \velfwd, \velang ) \in \reals^2$. 
Given a setpoint update $\action_{\idxtimestep} = (\velfwd_{\idxtimestep}, \velang_{\idxtimestep})$ at time $\idxtimestep$, we set the controller setpoint $\configuration = (\velfwd_{\idxtimestep}, \velang_{\idxtimestep})$ that is maintained between actions.
Given a start state $\state_1$ and an instruction $\instruction$, an execution $\execution$ of length $\execlen$ is a sequence  $\langle (\state_1, \action_1), \dots, (\state_{\execlen}, \action_{\execlen}) \rangle$, where $\state_\idxtimestep$ is the state at time $\idxtimestep$, $\action_{\idxtimestep<\execlen} \in \reals^2$ are setpoint updates, and $\action_{\execlen} = \stopaction$. 
The agent has  access to raw first-person monocular observations and pose estimates, and does not have access to the world state. The agent also has access  to an object database $\nod = \{\object^{(1)}, \ldots , \object^{(k)}\}$, where each object $\object^{(i)} =  (\{\query_{1}^{(i)}, \ldots , \query_{q}^{(i)}\}, \{\descr_{1}^{(i)}, \ldots, \descr_{w}^{(i)}\})$ is represented by sets of images $\query_{j}^{(i)}$ and natural language  descriptions $W_{j}^{(i)}$.
This database allows the agent to reason about previously unseen objects. 
At time $\idxtimestep$, the agent observes the \emph{agent context} $\context_{\idxtimestep} = (\instruction, \image_1,\ldots, \image_{\idxtimestep}, \pose_1, \ldots \pose_{\idxtimestep}, \nod)$,  where $\instruction$ is the instruction, $\image_i$ and $\pose_i$ are monocular first-person RGB images and 6-DOF agent poses observed at time $i = 1,\dots,t$, and $\nod$ is the object database.

\paragraph{Policy Model}

We use the two-stage policy decomposition of the Position Visitation Network v2~\cite[PVN2;][]{blukis2018following,blukis2019learning}: (a) predict the probability of visiting each position during instruction execution and (b) generate actions that visit high probability positions. 
We introduce a new method that uses an object database to identify references to objects in the instruction and segments these objects in the observed images (Section~\ref{sec:segment}). The instruction text is combined with the segmentation masks to create an object-centric map (Section~\ref{sec:objmap}), which is used as input to the two-stage policy (Section~\ref{sec:model}). 

\paragraph{Learning}

We train two language-conditioned object segmentation components, for simulated and physical environments (Section~\ref{sec:segment:learning}). For both, we use synthetically generated augmented reality training data using 3D objects overlaid on first-person environment images. 
We train our policy in simulation only,  using a demonstration dataset $\dataset^{\domainsim}$  that includes $\trainsetsize^{\domainsim}$ examples $\{ (\instruction^{(i)},  \execution^{(i)})\}_{i = 1}^{\trainsetsize^{\domainsim}}$, where $\instruction^{(i)}$ is an instruction, and $\execution^{(i)}$ is an execution (Section~\ref{sec:model}).
We use Supervised and Reinforcement Asynchronous Learning~\cite[$\rail$;][]{blukis2019learning}, an algorithm that concurrently trains the two model stages in two separate asynchronous processes. 
To deploy on the physical drone, we simply swap the object segmentation component with the one trained on real images. 

\paragraph{Evaluation}
We evaluate on a test set of $\testsetsize$ examples $\{ (\instruction^{(i)}, \state_1^{(i)},  \posseq^{(i)}) \}_{i = 1}^{\testsetsize}$, where $\instruction^{(i)}$ is an instruction, $\state_1^{(i)}$ is a start state, and $\posseq^{(i)}$ is a human demonstration. Test examples include only previously unseen instructions, environment layouts, trajectories, and objects.
We use human evaluation to verify if the generated trajectories are semantically  correct with regard to the instruction. 
We also use automated metrics. 
We consider the task successful if the agent stops  within a predefined Euclidean distance of the final position in $\posseq^{(i)}$. 
We evaluate the quality of the generated trajectory using earth mover's distance between $\posseq^{(i)}$ and executed trajectories.

\section{Few-shot Language-conditioned Segmentation}\label{sec:segment}

\begin{figure}[t]
\centering
\includegraphics[scale=0.34,clip,trim=0 1 12 16]{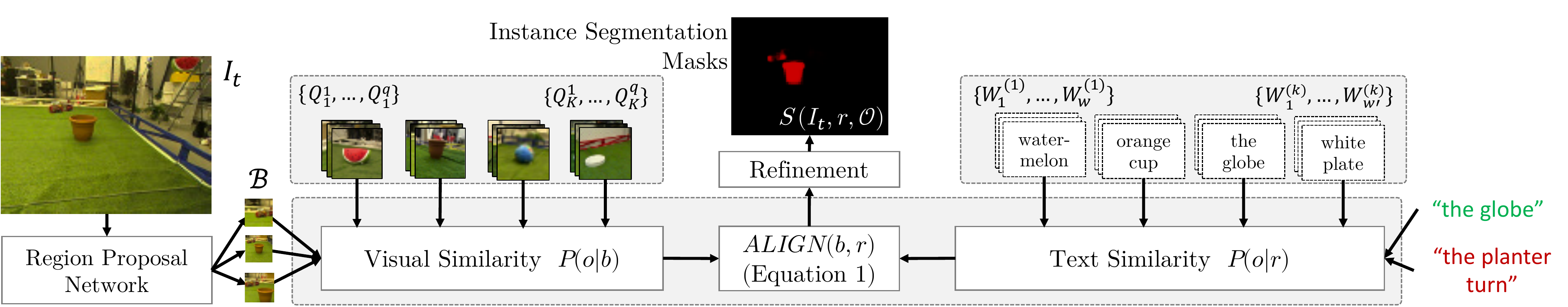}
\vspace{-19pt}
\caption{Few-shot language-conditioned segmentation illustration. Alignment scores are computed by comparing the visual similarity of database images to proposed bounding boxes and the textual similarity of database phrases with object references (e.g., the noisy ``the planter turn''). The aligned bounding boxes are refined to create segmentation masks for each mentioned object.}\label{fig:segmodel}
\end{figure}

Given a first-person RGB observation $\image$, a natural language phrase $\objref$, and an object database $\nod = \{\object^{(1)}, \ldots , \object^{(k)}\}$, the segmentation component  generates a first-person segmentation mask $\segmodel(\image, \objref; \nod)$ over $\image$. 
The segmentation mask has the same spatial dimensions as $\image$, and contains high values at pixels that overlap with the object referenced by $\objref$, and low values elsewhere.
The segmentation component additionally outputs an auxiliary  mask $\segmodel(\image)$ that assigns high values to pixels that overlap with \emph{any} object, and is primarily useful to learn collision-avoidance behaviors.
We use the output masks to generate object context maps (Section~\ref{sec:objmap}). 
Figure~\ref{fig:segmodel} illustrates the mask computation. 

\subsection{Segmentation Mask Computation}\label{sec:segment:align}

\newcommand{\alignscore}{\textsc{Align}}

We compute an alignment score between the phrase $\objref$ and proposed bounding boxes, and refine the bounding boxes to generate the pixel-wise mask. The alignment score combines several  probabilities:

\begin{small}
\begin{equation}
    \alignscore(\bbox, \objref) = \sum_{\object \in \nod}\hat{P}(\bbox  \mid  \object)\hat{P}(\object \mid \objref) 
    = \sum_{\object \in \nod}\frac{\hat{P}(\object \mid \bbox)\hat{P}(\bbox)\hat{P}(\object \mid \objref)}{\hat{P}(\object)}\;\;,\label{eq:grounding-inference}
\end{equation}
\end{small}

\noindent
where $\bbox$ is a bounding box and the second equality is computed using Bayes' rule. 
The use of distributions allows us to easily combine separate quantities while controlling for their magnitude and sign. 
We assume $\hat{P}(\object)$ to be uniform, and compute each of the other three quantities separately. 

We use a region proposal network ($\rpn$) to produce bounding boxes $\bboxset = \{\bbox^{(j)}\}_j$ over the image $\image$. Each $\bbox^{(j)}$ corresponds to a region $\image[\bbox^{(j)}]$ in $\image$  likely to contain an object. 
$\rpn$ also computes the probability that each bounding box contains an object $P(\image[\bbox^{(j)}] \texttt{ is an object})$, which we use for $\hat{P}(\bbox)$ in Equation~\ref{eq:grounding-inference} with the assumption that these quantities are proportional.
We use the RPN implementation from the Detectron2 object recognizer~\cite{wu2019detectron2}.

We estimate the probability $\hat{P}(\object \mid \bbox)$ that an object $\object$ appears in a proposed image region $\image[\bbox]$ using visual similarity. 
Each object $\object \in \nod$ is associated with a small set of images $\{\query_{1}, \ldots , \query_{q}\}$. We compute the similarity between these images and $\image[\bbox]$.  
We use $\imageembedding$ to map each $\query_{j}$ and the image region $\image[\bbox]$  to a vector representation. 
$\imageembedding$ is a convolutional neural network (CNN) that maps each image to a metric space where the L2 distance captures visual similarity.
We estimate a probability density $\pdf(\bbox \mid \object)$ using Kernel Density Estimation with a symmetrical multivariate Gaussian kernel.
The probability is computed with Bayes' rule and normalization:

\begin{small}
\begin{equation}
  \hat{P}(\object \mid \bbox) =
  \frac{pdf(\object \mid \bbox)}{\sum_{\object' \in \nod}pdf(\object' \mid \bbox)} =
  \frac{pdf(\bbox \mid \object)\hat{P}(\object)/\hat{P}(\bbox)}{\sum_{\object' \in \nod}pdf(\bbox \mid \object')\hat{P}(\object')/\hat{P}(\bbox)} =
  \frac{pdf(\bbox \mid \object)\hat{P}(\object)}{\sum_{\object' \in \nod}pdf(\bbox \mid \object')\hat{P}(\object')}
  \;\;.
\end{equation}
\end{small}

\noindent
We compute $\hat{P}(\object \mid \objref)$ using the same method as $\hat{P}(\object \mid \bbox)$. 
The phrases $\{\descr_{1}, \ldots, \descr_{w}\}$ associated with the object $\object$ take the place of associated images, and the phrase $\objref$ is used in the same role as the image region. 
We compute phrase embeddings as the mean of \textsc{GloVe} word embeddings~\cite{pennington2014glove}.

While we can use $\alignscore$ to compute the mask, it only captures the  square outline of objects. 
We refine each box into a segmentation mask that follows the contours of the bounded object. The masks and alignment scores are used to compute the mask for the object mentioned in the phrase $\objref$.  
We use a U-Net~\cite{ronneberger2015u} architecture to map each image region $\image[\bbox]$ to a mask $\textbf{M}_b$ of the same size as $\bbox$. The value $[\mathbf{M}_b]_{(x,y)}$ is the probability that it belongs to the most prominent object in the region.

The first-person object segmentation mask value $\segmodel(\image, \objref; \nod)$  for each pixel $(x,y)$ is  the sum of segmentation masks from all image regions $\bboxset$, weighed by the probability that the region contains $\objref$:

\begin{small}
    \begin{equation}
        {[\segmodel(\image, \objref; \nod)]}_{(x,y)} = \sum_{\bbox \in \bboxset}\alignscore(\bbox, \objref){[\mathbf{M}_b]}_{(x,y)}\;\;,
    \end{equation}
\end{small}

\noindent
where $\alignscore(\cdot)$ is defined in Equation~\ref{eq:grounding-inference}.
The auxiliary segmentation mask of all objects  $\segmodel(\image)$, including unmentioned ones, is $[\segmodel(\image)]_{(x,y)} = \max_{\bbox \in \bboxset}[\mathbf{M}_b]_{(x,y)}$.

\subsection{Learning}\label{sec:segment:learning}

Learning the segmentation function $\segmodel(\image, \objref; \nod)$ includes estimating the parameters of the image embedding $\imageembedding$, the region proposal network $\rpn$, and the refinement U-Net.
We use pre-trained \textsc{GloVe} embeddings to represent object references $\objref$.
We train with a dataset $\objectdata = \{(\image^{(i)},\{ (\bbox^{(i)}_j, \objmask^{(i)}_j, \object^{(i)}_j) \}_j)\}_{i}$, where $\image^{(i)}$ is a first-person image and $\bbox^{(i)}_j$ is a bounding box of the object $\object^{(i)}_j$, which has the mask $\objmask^{(i)}_j$. 
We generate $\objectdata$ by overlaying 3D objects on images from the physical or simulated environments. Appendix~\ref{app:ardata} describes this process. 
Using a large number of diverse objects allows to generalize beyond specific object classes to support new, previously unseen objects at test-time.
We train the $\rpn$ from scratch using the Detectron2 method~\cite{wu2019detectron2} and $\objectdata$.

We use image similarity metric learning to train $\imageembedding$. We extract triplets $\{(\image_{a}^{i}, \{\image_{a'}^{ij}\}_{j}, \{\image_{b}^{ij}\}_{j})\}_{i}$ from the object dataset $\objectdata$. Each object image $\image_{a}^{i}$ is coupled with  images $\{\image_{a'}^{ij}\}_{j}$ of the same object, and images $\{\image_{b}^{ij}\}_{j}$ of a randomly drawn different object.
Images include varying  lighting conditions and viewing angles.
We train $\imageembedding$ by optimizing a max-margin triplet loss $\tripletloss$:

\begin{small}
    \begin{equation}
        \tripletloss(\image_{a}^{i}, \{\image_{a'}^{ij}\}_{j}, \{\image_{b}^{ij}\}_{j}) = \max(s_{a} - \tripletmarginB, 0)
        + \max(-s_{b} + \tripletmarginB, 0) + \max(s_{a} - s_{b} + \tripletmargin, 0)\label{eq:triplet}
    \end{equation}
\end{small}

\begin{small}
\begin{equation*}
    s_{a} = \min_{j}\lvert\imageembedding(\image_{a}^{i}) - \imageembedding(\image_{a'}^{ij})\rvert_{2}^{2} \quad\quad
    s_{b} = \min_{j}\lvert\imageembedding(\image_{a}^{i}) - \imageembedding(\image_{b}^{ij})\rvert_{2}^{2}\;\;.
\end{equation*}
\end{small}

\noindent
$\tripletmargin$ and $\tripletmarginB$ are margin constants. $s_{a}$ and $s_{b}$ are distances between an image and a set of images.
The first term in Equation~\ref{eq:triplet} encourages images of the same object to be within a distance of at most $\tripletmarginB$ of each other. The second term pushes images of different objects to be at least $\tripletmarginB$ far from each other. 
The third term encourages the distance between images of the same object to be  smaller than between images of different objects by at least  $\tripletmargin$.

We train the refinement U-Net with  data $\{(\image^{(i)}[\bbox_{j}], \objmask^{(i)}_{j})\}_{i}$ of $\image^{(i)}[\bbox_{j}]$ image regions and $\objmask^{(i)}_{j}$ zero-one valued ground truth masks generated from $\objectdata$. We use a  pixel-wise binary cross-entropy loss.

\section{Object Context Grounding Maps}\label{sec:objmap}

We compute  an allocentric \emph{object context grounding map} of the world that combines (a) information about object locations from the segmentation component (Section~\ref{sec:segment}) and (b) information about how to interact with objects, which is derived from the language context around object mentions in the instruction $\instruction$. 
The map is created from a sequence of observations. 
At timestep $\idxtimestep$, we denote the map  $\contextmapworld_\idxtimestep$.
Constructing $\contextmapworld_\idxtimestep$ involves identifying and aligning text mentions and observations of objects using language-conditioned segmentation, accumulating over time the segmentation masks projected to an allocentric reference frame, and encoding the language context of object mentions in the map. 
This process is integrated into the first stage of our policy (Section~\ref{sec:model}), and illustrated in Figure~\ref{fig:model}. 

\paragraph{Language Representation}

Given the instruction $\instruction$, we generate (a) a multiset of object references  $\objrefset$, (b) contextualized representation $\contextemb(\objref)$ for each $\objref \in \objrefset$, and (c)  an object-independent instruction representation $\mathbf{\hat{h}}$. 
The set of object references $\objrefset$ from $\instruction$ is $\{\objref \mid   \objref \in \textsc{Chunker}(\instruction) \land   \objrefclassifier(\objref, \nod)\}$, where $\textsc{Chunker}$ is a noun phrase chunker~\cite{Tjong2000:chunking} and $\objrefclassifier$ is an object reference boolean classifier. 
For example, $\textsc{Chunker}$ may extract \nlstring{the globe} or \nlstring{it}, and $\objrefclassifier$ will only classify \nlstring{the globe} as an object reference. We use the pre-trained spaCy chunker~\cite{spacy2}, and train a two-layer fully connected neural network classifier for $\objrefclassifier$. Appendix~\ref{app:model:objref} provides more details.  

We remove all object references from $\instruction$ to create $\anoninstruction = \langle \hat{u}_0, \ldots, \hat{u}_l \rangle$ by replacing all object reference spans with the placeholder token $\objreftok$.
$\anoninstruction$ is a sequence of tokens that captures aspects of navigation behavior, such as trajectory shape and spatial relations that do not pertain to object identities and would generalize to new objects.
We encode $\anoninstruction$ with a bi-directional long short-term memory~\cite[LSTM;][]{hochreiter1997long}  recurrent neural entwork (RNN) to generate a sequence of hidden states $\langle\mathbf{h}_1,\ldots,\mathbf{h}_l\rangle$. The contextualized representation $\contextemb(\objref)$ for each object reference $\objref$ is $\mathbf{h}_i$ for the placeholder token replacing it. $\contextemb(\objref)$ captures contextual information about the object within the instruction, but does not contain information about the object reference itself. We define the object-independent instruction representation as $\mathbf{\hat{h}} = \frac{1}{l}\sum_{i = 1}^l\mathbf{h}_i$. 

We train $\objrefclassifier$  using an object reference dataset of noun chunks  labeled to indicate whether they are referring to physical objects.
Appendix~\ref{app:alignments} describes a general technique for automatically generating this data from any navigation dataset that includes instructions, ground-truth trajectories, and object position annotations (e.g., \textsc{Room2Room}~\cite{anderson2017vision}, Lani~\cite{misra2018mapping}). The language representation $\contextemb$ is trained end-to-end with the complete instruction-following policy (Section~\ref{sec:model}). 

\paragraph{Object Context Mapping}

At each timestep $\idxtimestep$, we compute the language-conditioned segmentation mask $\segmodel(\image_\idxtimestep, \objref, \nod)$ that identifies each object $\objref \in \objrefset$ in the first-person image $\image_\idxtimestep$, and the all-object mask $\segmodel(\image_\idxtimestep)$ that identifies all objects  (Section~\ref{sec:segment}).

We use differentiable geometric operations to construct an allocentric object context grounding map $\contextmapworld_\idxtimestep$. Each position in the environment is represented with a learned vector that encodes whether it contains an object, if the contained object was mentioned in the instruction, and the instruction context of the object mention (e.g., whether the agent should pass it on its left or right side).
The map encodes desired behavior with relation to objects, but abstracts away object identities and properties.

We project the set of masks $\{\segmodel(\image_\idxtimestep, \objref, \nod) \mid \objref \in \objrefset\}\cup\{\segmodel(\image_\idxtimestep)\}$ to an allocentric world reference frame using a pinhole camera model to obtain a set of  allocentric object segmentation masks  that identify each object's location in the global environment coordinates.
We accumulate the projected masks over time by computing the max across all previous timesteps for each position to compute allocentric masks:
$\{\segmaskworld(\image_\idxtimestep, \objref, \nod) \mid \objref \in \objrefset\}\cup\{\segmaskworld(\image_\idxtimestep)\}$.
We combine the object reference contextualized representations with the masks to compute an object context grounding map $\contextmapworld_\idxtimestep$:

\begin{small}
\begin{equation}
    \contextmapworld_\idxtimestep = [\sum_{\objref \in \objrefset}\contextemb(\objref) \cdot \segmaskworld(\image_\idxtimestep, \objref, \nod); \segmaskworld(\image_\idxtimestep); \boundaryworld_\idxtimestep] \;\;, 
\end{equation}
\end{small}

\noindent
where $\boundaryworld_\idxtimestep$ is a 0/1 valued mask indicating environment boundaries, and $[\cdot ; \cdot]$ is a channel-wise concatenation. The product $\contextemb(\objref) \cdot \segmaskworld(\image_\idxtimestep, \objref, \nod)$ places the contextualized object reference representation for $\objref$ in the environment positions containing objects aligned to it. The summation across all $\objrefset$ creates a single tensor of spatially-placed contextualized representations.

\section{Integration into an Instruction-following Policy}\label{sec:model}

\begin{figure*}[t]
\centering
\includegraphics[scale=0.24]{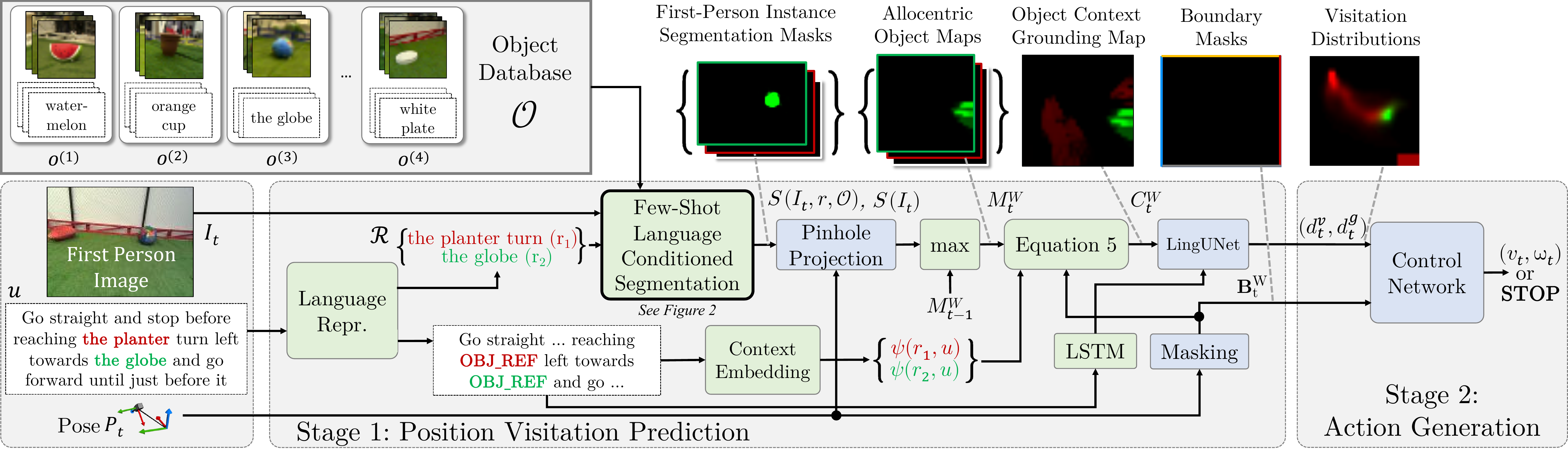}
\vspace{-8pt}
\caption{Policy architecture illustration. The first stage uses our few-shot language-conditioned segmentation to identify mentioned objects in the image. The segmentation and instruction embedding are used to generate an allocentric object context grounding map $\contextmapworld_\idxtimestep$, a learned map of the environment that encodes at every position the behavior to be performed at or near it. We use $\lingunet$ to predict visitation distributions, which the second stage maps to velocity commands. 
The components in blue are adopted from prior work~\cite{blukis2018following, blukis2019learning}, while we add the components in green to enable few-shot generalization. 
Appendix~\ref{app:model} includes a whole-page version of this figure.
}\label{fig:model}
\end{figure*}

We integrate the language-conditioned segmentation model $\segmodel$ and the object context grounding map $\contextmapworld$ with an existing representation learning instruction-following policy to allow it to reason about previously unseen objects.
We use the Position Visitation Network~\cite{blukis2018following, blukis2019learning}, but our approach is applicable to other policy architectures~\cite{anderson2019chasing, krantz2020beyond}.

Given the agent context $\context_\idxtimestep$ at time $\idxtimestep$, the policy $\pi$ outputs the $\stopaction$ action probability $\stopprob_\idxtimestep$, a forward velocity $\velfwd_{\idxtimestep}$, and an angular velocity $\velang_{\idxtimestep}$.
The policy model $\pi(\context_{\idxtimestep}) = \stageB(\stageAreal(\context_{\idxtimestep}))$ decomposes to  two stages. 
The first stage $\stageAreal$ predicts two \emph{visitation distributions} over environment positions: a trajectory distribution $\trajvisit$ indicating the probability of passing through a position and a goal distribution $\stopvisit$ giving the probability of the $\stopaction$ action at a position.
The second stage $\stageB$ outputs velocity commands or $\stopaction$ to create a trajectory that follows the distributions by visiting high probability positions according to $\trajvisit$, and stopping in a likely position according to $\stopvisit$. 
Figure~\ref{fig:model} illustrates the model. 
We integrate our few-shot segmentation (Section~\ref{sec:segment}) and mapping (Section~\ref{sec:objmap}) into the first stage $\stageAreal$.
Following previous work~\cite{misra2018mapping,blukis2018mapping,blukis2019learning}, we use the $\lingunet$ architecture to predict the visitation distributions $\trajvisit_t$ and $\stopvisit_t$. 
Appendix~\ref{app:model:lingunet} reviews $\lingunet$. 
We use the object context map $\contextmapworld_\idxtimestep$ and the object-independent instruction representation vector $\mathbf{\hat{h}}$ as inputs to $\lingunet$. Both are conditioned on the the object database $\nod$ used for language-conditioned segmentation, and designed to be otherwise indifferent to the visual or semantic properties of specific objects. 
This makes the policy easy to extend to reason about previously unseen objects by simply adding them to the database.
In contrast, \citet{blukis2019learning} uses an embedding of the full instruction and learned semantic and grounding maps as input to $\lingunet$. 
These inputs are trained to reason about a fixed set of objects in images and text, and do not generalize to new objects, as demonstrated by our experiments (Section~\ref{sec:results}). 
We use the second stage $\stageB$ control network design of \citet{blukis2019learning} (Appendix~\ref{app:model:control-network}).

\paragraph{Policy Training}

We use \fullsureal~\cite[$\sureal$;][]{blukis2019learning} to estimate the parameters $\paramsA$ for the first stage $\stageA(\cdot)$ and $\paramsB$ for the second stage $\stageB(\cdot)$. 
In contrast to \citet{blukis2019learning}, we do not use a domain-adversarial loss to jointly learn for both the simulation and physical environment.
Instead, we train two separate language-conditioned segmentation models, one for training in simulation, and one for  testing on the physical agent.
This does not require a significant change to the training process. 
Roughly speaking, $\sureal$ trains the two stages concurrently in two  processes. A supervised learning process is used to train the first stage, and a reinforcement learning process for the second. 
The processes constantly exchange information so the two stages work well together. Appendix~\ref{app:learning} describes  $\sureal$ and the loss terms. 
Deployment on the real robot after training in simulation requires only swapping the segmentation model, and does not require any targeted domain randomization beyond the randomness inherent in the AR training data.

\section{Experimental Setup}\label{sec:experiments}

\paragraph{Environment and Data}
We use the physical environment and data of \citet{blukis2019learning} (Figure~\ref{fig:intro}), and expand it with new objects. 
We use the quadcopter simulator of \citet{blukis2018following}.
We use 41,508 instruction-demonstration training pairs from \citet{blukis2019learning} for training. 
We collect additional data with eight new, previously unseen objects for testing our method and training the $\pvntwoall$ baseline.
Appendix~\ref{app:data:corpora} provides complete details, including the set of new objects. 
The data contains one-segment and longer two-segment instructions. We use both for training, but only evaluate with the more complex two-segment data. 
For evaluation in the physical environment, we use 63 instructions with new objects or 73 with seen objects. 
We use a fixed object database with all unseen objects at test time. 
It contains five images and five phrases per object. 
Appendix~\ref{app:objectdata} provides additional details and the complete database. 
We generate language-conditioned segmentation training data (Section~\ref{sec:segment:learning}) by collecting random flight trajectories in empty physical and simulation environments, and using augmented reality to instantiate randomly placed ShapeNet~\cite{chang2015shapenet} objects with automatically generated bounding box and segmentation mask annotations. 
Appendix~\ref{app:ardata} shows examples.

\paragraph{Evaluation}

We follow the evaluation setup of \citet{blukis2019learning}. We use human evaluation on Amazon Mechanical Turk using top-down animations to score  the agent's final stopping position (goal score) and the complete trajectory (path score), both judged in terms of adhering to the instruction using a five-point Likert score.
We also report: (a) \textsc{SR}: success rate of stopping within 47cm of the demonstration stopping position; and (b) \textsc{EMD}: earth mover's distance in meters between the agent and demonstration trajectories.

\paragraph{Systems}
We train our approach $\fspvn$ on the original training data and compare it to two versions of $\pvntwo$~\cite{blukis2018mapping}, the previous state of the art on this task:
(a) $\pvntwoall$: the $\pvntwo$ model trained on all training data, including all new objects; 
(b) $\pvntwoseen$: the $\pvntwo$ model trained only on the original training data, the same data we use with our model.
$\pvntwo$ is not designed to generalize to new objects, as $\pvntwoseen$ shows. 
To correctly deploy $\pvntwo$ in a new environment, it has to be trained on large amount of instruction data that includes the new objects, which is reflected in $\pvntwoall$ that encounters the new objects hundreds of times during training. In contrast, our model only has access to a small object database $\nod$ that can be quickly constructed by an end user.
We also report two non-learning systems:
(a) $\textsc{Average}$: outputs average training data velocities for the average number of steps; 
(b) $\textsc{Oracle}$: a hand-crafted upper-bound expert policy that has access to the ground-truth demonstration.
Appendix~\ref{app:impl} provides implementation details.

\section{Results}\label{sec:results}

\begin{figure}[t]
    \centering
    \includegraphics[scale=0.25,clip,trim=4 0 25 0]{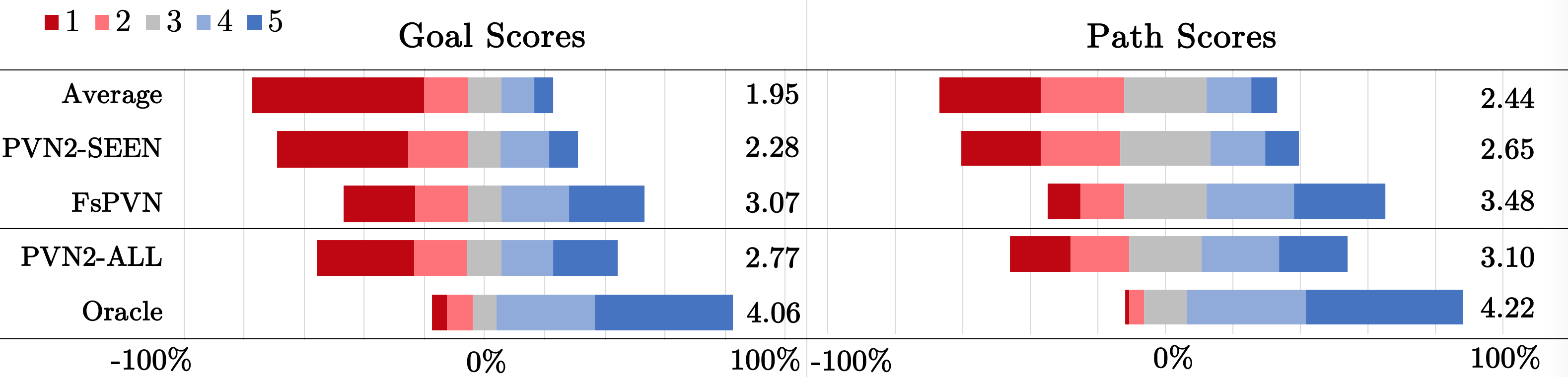}
    \vspace{-5pt}
    \caption{Human evaluation results on the physical quadcopter in environments with only new objects. We plot the Likert scores using Gantt charts of score frequencies, with mean scores in black.}\label{fig:human_results}
\end{figure}

Figure~\ref{fig:human_results} shows human evaluation Likert scores on the physical environment. 
A score of 4--5 reflects good performance. 
$\fspvn$ receives good scores 47\% of the time for correctly reaching the specified goal, and 53\% of the time for following the correct path, a significant improvement over $\pvntwoseen$. 
This shows effective generalization to handling new objects. 
$\fspvn$ outperforms $\pvntwoall$ even though the former has seen all objects during training, potentially because the object-centric inductive bias simplifies the learning problem. 
The imperfect $\sysoracle$ performance highlights the inherent ambiguity and subjectivity of natural language instruction. 

\begin{table}[t]
\vspace{+8pt}
\setlength{\tabcolsep}{4pt}
\begin{minipage}{.65\textwidth}
\centering
\footnotesize
\begin{tabular}{@{}l c c c c c c@{}}
\toprule
\multirow{2}{*}{Method}  & \multicolumn{2}{c}{Physical Env.} &\multicolumn{2}{c}{Simulation}   & \multicolumn{2}{c@{}}{Simulation}\\
& SR $\uparrow$  & EMD $\downarrow$  & SR $\uparrow$ & EMD $\downarrow$ & SR $\uparrow$ & EMD $\downarrow$\\
\midrule
\multicolumn{1}{@{}l}{\textbf{Test Results}} & \multicolumn{4}{l}{\textbf{w/8 New Objects}} &  \multicolumn{2}{l@{}}{\textbf{w/15 Seen Objects}}\vspace{2pt} \\
\cmidrule(lr){2-5} 
\cmidrule(l){6-7}
\avgmodel              & 12.7   & 0.63  & 15.9   & 0.70   & 13.7   & 0.78  \\
\textsc{$\pvntwoseen$}  & \phantom{0}3.2   & 0.65  & 27.0   & 0.59     & 43.8   & 0.60 \\
\textsc{$\fspvn$} & \textbf{28.6}   & \textbf{0.45}  & \textbf{34.9} & \textbf{0.42}    & 46.6   & 0.48  \\
\midrule[0.1pt]
\textsc{$\pvntwoall$} & 30.2   & 0.49  & 49.2   & 0.40  & 37.0   & 0.53  \\
\oraclemodel          & 95.2   & 0.22  & 98.4   & 0.16  & 97.3   & 0.17  \\
\bottomrule
\end{tabular}
\end{minipage}~
\begin{minipage}{.25\textwidth}
\centering
\footnotesize
\begin{tabular}{@{}l c c@{}}
\toprule
\multirow{2}{*}{Method}   &\multicolumn{2}{c@{}}{Simulation}\\
& SR $\uparrow$ & EMD $\downarrow$ \\
\midrule
\multicolumn{3}{@{}l@{}}{\textbf{Dev. Results w/8 New Objects}}\vspace{11pt} \\
\textsc{$\fspvn$}  & 28.2   & 0.52      \\
\textsc{$\fspvnBC$}  & 20.4   & 0.68      \\
\textsc{$\fspvnbignod$}  & 27.2   & 0.52  \\
\textsc{$\fspvndeaf$}  & 12.6   & 0.70    \\
\textsc{$\fspvnblind$}  & 15.5   & 0.58   \\

\bottomrule
\end{tabular}
\end{minipage}
\caption{Automated evaluation test (left) and development (right) results.  SR:\@ success rate (\%) and EMD:\@ earth-mover's distance in meters between agent and demonstration trajectories.
}\label{tab:auto_results}
\vspace{-10pt}
\end{table}

Unlike $\textsc{PVN2}$, our approach learns instruction following behavior entirely in simulation, and utilizes a separately trained few-shot segmentation component to deploy in the real world. As a result, the simulation no longer needs to include the same objects as in the real world. This removes an important bottleneck of scaling the simulator towards real-world applications. Additionally, $\textsc{PVN2}$ uses  auxiliary objectives that require object and identity information during training. $\fspvn$ does not use these, and does not require object-level annotation in the instruction training data.

Table~\ref{tab:auto_results} (left) shows the automated metrics. \textsc{EMD} is the more reliable metric of the two because it considers the entire trajectory. 
$\fspvn$ is competitive to $\pvntwoall$ in the physical environment on previously unseen objects. $\pvntwoall$ slightly outperforms our approach according to the automated metrics, contrary to human judgements.
This could be explained by $\fspvn$ occasionally favoring trajectories that are semantically correct, but differ from demonstration data.
$\pvntwoseen$ performs significantly worse, with only 3.2\% SR and 0.59 EMD on unseen objects. 
We observe that it frequently explores the environment endlessly, never gaining confidence that it has observed the goal.
$\pvntwoseen$ performs much better in simulation,  potentially because it encounters more objects in simulation, which allows it to learn to focus on properties (e.g., colors) that are also used with new objects. 
Comparing simulation performance between previously unseen and seen objects, we observe that even though our approach generalizes well to unseen objects, there remains a performance gap.

Table~\ref{tab:auto_results} (right) shows ablation results. 
$\fspvnbignod$ is the same model as $\fspvn$, but uses a larger object database including 71 objects during test time. This significant increase in database size leads to a modest decrease in performance. 
$\fspvnBC$ replaces $\sureal$ with behavior cloning, illustrating the benefit of of exploration during training. 
We study two sensory-inhibited ablations that perform poorly:  $\fspvnblind$ receives a blank image and $\fspvndeaf$ an empty instruction. 

Finally, Appendix~\ref{app:results} provides an evaluation of our language-conditioned segmentation methods and image similarity measure in isolation. 
Our approach offers the benefit of interpretable object grounding via the recovered alignments. Appendix~\ref{app:demo} provides example alignments.

\section{Conclusion}\label{sec:conclusion}

We focus on the problem of extending a representation learning instruction-following model to reason about new objects, including their mentions in natural language instructions and observations in raw images. 
We propose a few-shot language-conditioned segmentation method, and show how to train it from easily generated synthetic data. 
This method recovers alignments between object mentions and observations, which we use to create an object-centric environment map that encodes how objects are used in a natural language instruction. 
This map forms an effective intermediate representation within a policy that maps natural language and raw observations to continuous control of a quadcopter drone. In contrast to previous learning methods, the robot system can be easily extended to reason about new objects by providing it with a small set of exemplars. 
It also offers the benefits of portability between simulation and real world and interpretability of object grounding via the recovered alignments. 
Our few-shot language-conditioned segmentation component is applicable to other tasks, including potentially on different robotic agents and other vision and language tasks.

\section*{Acknowledgments}

This research was supported by a Google Focused Award and NSF CAREER-1750499. We thank Ge Gao, Noriyuki Kojima, Alane Suhr, and the anonymous reviewers for their helpful comments.

\bibliography{references}

\clearpage

\appendix

\section{Internal Model Reasoning Visualization}\label{app:demo}

\begin{figure}[t!]
\centering
\includegraphics[scale=0.24]{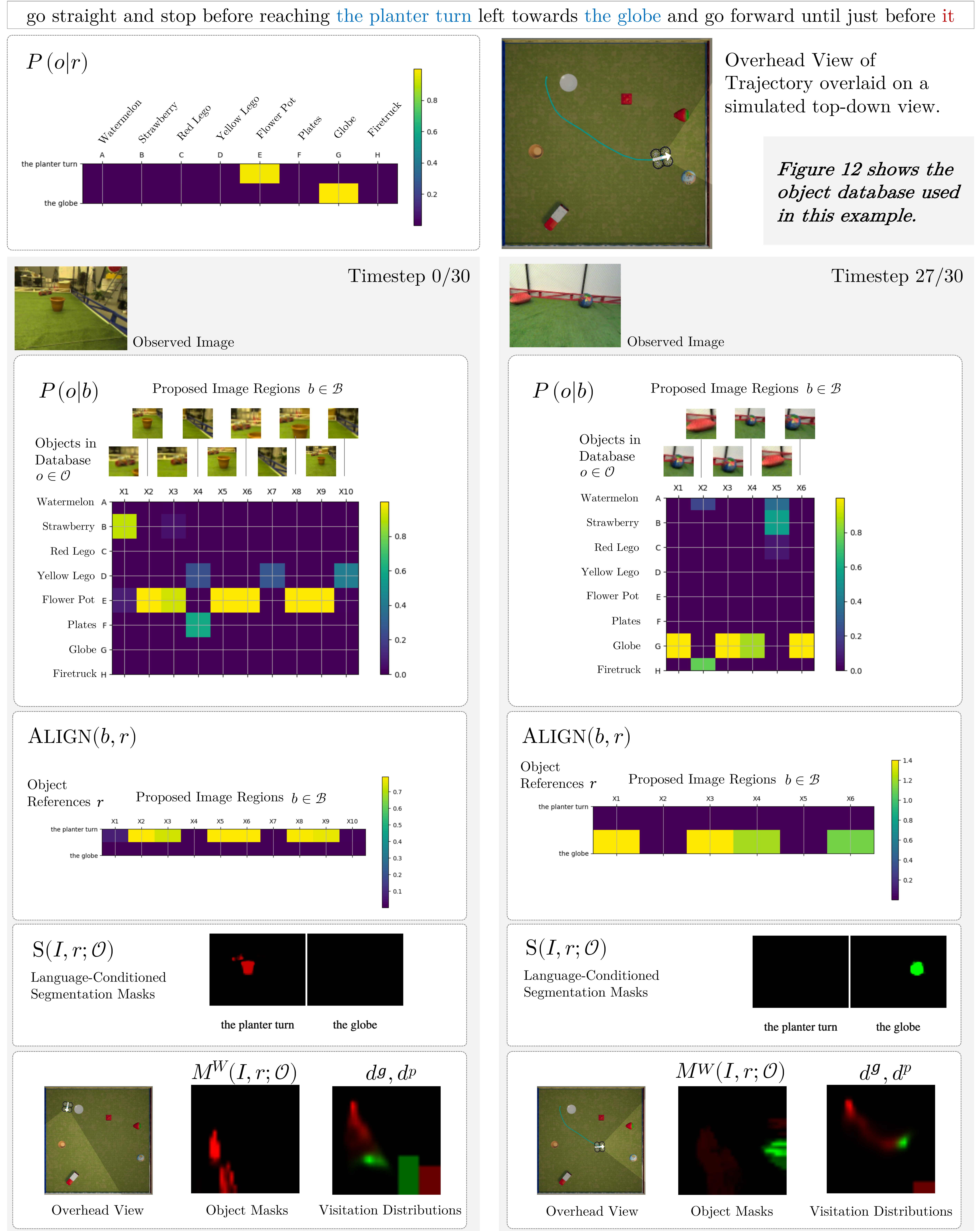}
\caption{Visualization of the model reasoning when executing the instruction \nlstring{go straight and stop before reaching the planter turn left towards the globe and go forward until just before it}. 
The extracted object references are highlighted in the instruction in blue, and other noun chunks in red. 
The probability $\hat{P}(o | r)$ that aligns each object reference with an object in the database is visualized at the top-left pane. An overhead view of the quadcopter trajectory visualized over a simulated image of the environment layout is given at the top-right pane.
For timestep 0 (left) and 27 (right), we show the first-person image $I_t$ observed at timestep $t$, the probability $\hat{P}(o|b)$ that aligns each proposed image region $b \in \mathcal{B}$ with an object in the database, the alignment score $\textsc{Align}(b,r)$ between image regions and object references computed from Equation~\ref{eq:grounding-inference}, the resulting first-person segmentation masks $S(\image,\objref,\nod)$, the projected object masks $M^W(I,r,\nod)$ obtained by projecting $S(\image,\objref,\nod)$ into an allocentric reference frame, and the predicted visitation distributions $\trajvisit$ (red) and $\stopvisit$ (green).}
\label{fig:app:demo}
\end{figure}

Figure~\ref{fig:app:demo} illustrates how the model reasoning when following an instruction can be visualized.

\section{Model Details}\label{app:model}

\begin{sidewaysfigure*}[p]
\centering
\includegraphics[scale=0.4]{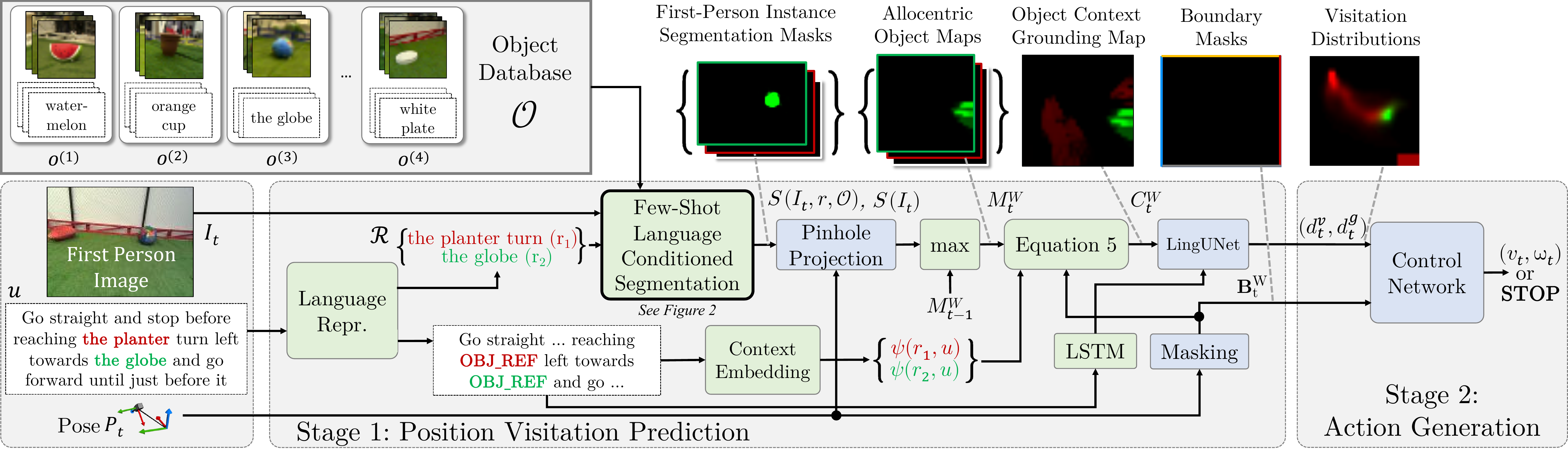}
\vspace{-8pt}
\caption{Full-page version of Figure~\ref{fig:model}. 
Policy architecture illustration. The first stage uses our few-shot language-conditioned segmentation to identify mentioned objects in the image. The segmentation and instruction embedding are used to generate an allocentric object context grounding map $\contextmapworld_\idxtimestep$, a learned map of the environment that encodes at every position the behavior to be performed at or near it. We use $\lingunet$ to predict visitation distributions, which the second stage maps to velocity commands. 
The components in blue are adopted from prior work~\cite{blukis2018following, blukis2019learning}, while we add the components in green to enable few-shot generalization.}\label{fig:model_fullpage}
\end{sidewaysfigure*}

Figure~\ref{fig:model_fullpage} shows a whole-page version of Figure~\ref{fig:model} from the main paper.

\subsection{Object Reference Classifier $\textsc{ObjRef}$}\label{app:model:objref}

\newcommand{\objrefnetwork}{\textsc{ObjRefNN}}
\newcommand{\glove}{\textsc{GloVe}}

The object reference classifier $\objrefclassifier$ inputs are a sequence of tokens representing a noun chunk and the object database $\nod$. The output is $\textsc{True}$ if the noun chunk refers to a physical object, and $\textsc{False}$ otherwise.

We represent noun chunks with pre-trained $\glove$ vectors~\cite{pennington2014glove}. We use the \texttt{en\_core\_web\_lg} model from the SpaCy library~\cite{spacy2}.\footnote{\url{https://spacy.io/}}
The classifier considers a noun chunk an object reference if either (a) a neural network object reference classifier assigns it a high score, or (b) the noun chunk is substantively similar to a phrase in the object database $\nod$. 
The classifier decision rule for a noun chunk $\hat{r}$ is:

\begin{small}
\begin{equation}
    \objrefclassifier(\hat{r}, \nod) = \objrefnetwork(\glove(\hat{r})) + \lambda_{R1} \min_{o^{(i)} \in \nod}\min_{j} \|\glove(\hat{r}) - \glove(\query_j^{(i)})\|_2^2 < T_{R2} \; ,
\end{equation}
\end{small}

\noindent
where $\objrefnetwork$ is a two-layer fully connected neural network, $\glove$ is a function that represents a phrase as the average of its token $\glove$ embeddings, $\lambda_{R1}$ is a hyperparameter that balances between the database-agnostic network $\objrefnetwork$ and similarity to the object database $\nod$, and $T_{R2}$ is a hyperparameter that adjusts precision and recall.

The classifier is trained on a dataset $\mathcal{D}_{R} = \{(\hat{r}^{(k)}, l^{(k)})\}_k$ of noun chunks paired with labels $l^{(k)} \in \{0, 1\}$ indicating whether the noun chunk is an object reference.
The procedure for extracting this data from a navigation instruction dataset is described in Appendix~\ref{app:alignments}.

\subsection{Contextualized Object Representations}\label{app:model:ctx}

\begin{figure}
\centering
\includegraphics[scale=0.4]{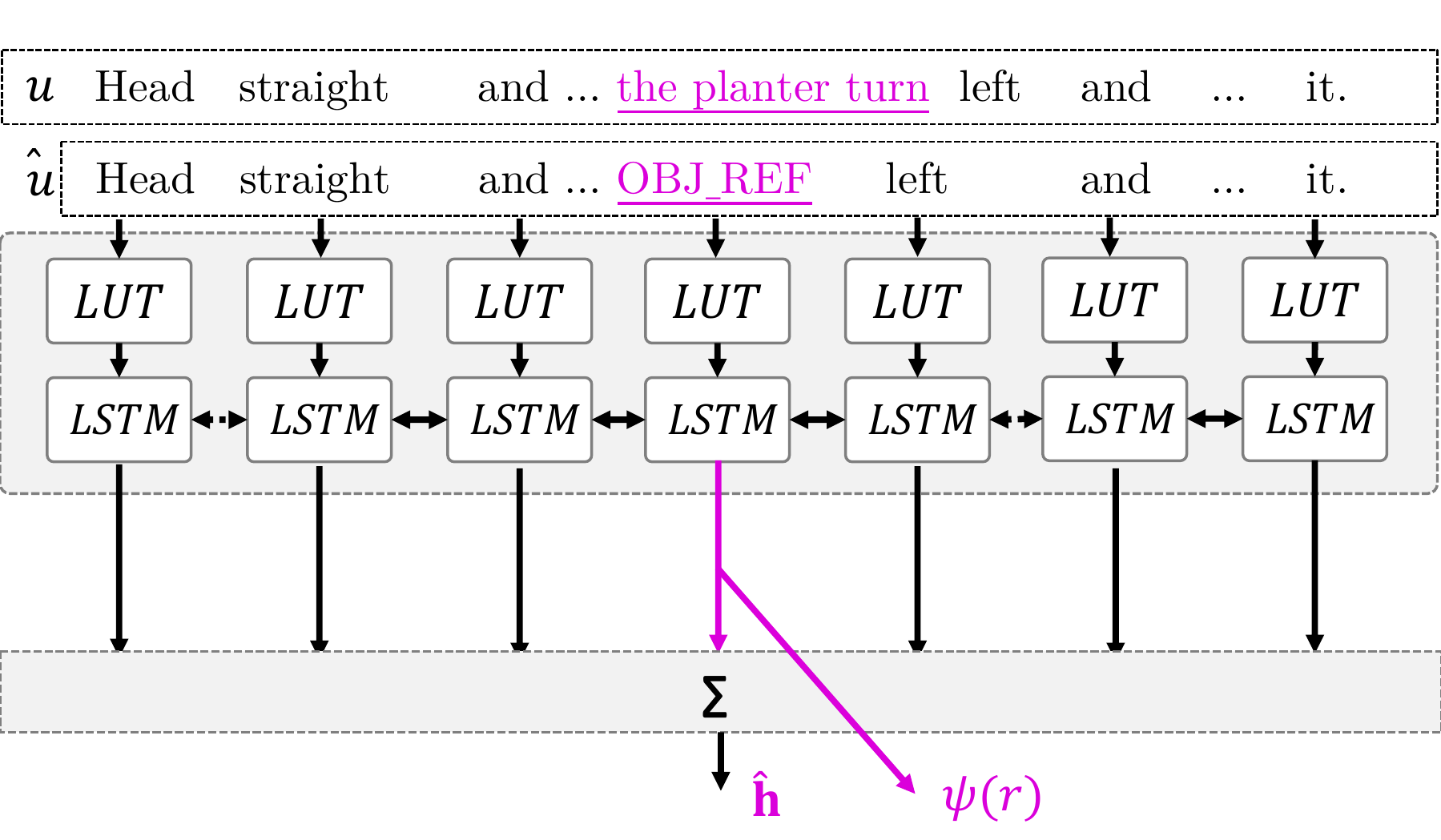}
\caption{Context embedding illustration. On top is a (shortened) instruction $\instruction$. The second row shows the corresponding anonymized instruction $\anoninstruction$. In the third row, we represent each word in $\anoninstruction$ with a vector from a look-up table (LUT), and then encode the sequence with a bi-directional LSTM.
The hidden states at positions corresponding to object reference tokens are object reference context embeddings. The sum of all hidden states is the anonymized instruction representation.}
\label{fig:contextembedding}
\end{figure}

Figure~\ref{fig:contextembedding} illustrates the neural network architecture of $\contextemb$ and the anonymized instruction representation $\mathbf{\hat{h}}$, where all objects mentions are replaced with placeholder tokens.

\subsubsection{LingUNet Computation for Predicting Visitation Distributions}\label{app:model:lingunet}

\begin{figure}[t]
\centering
\includegraphics[scale=0.4]{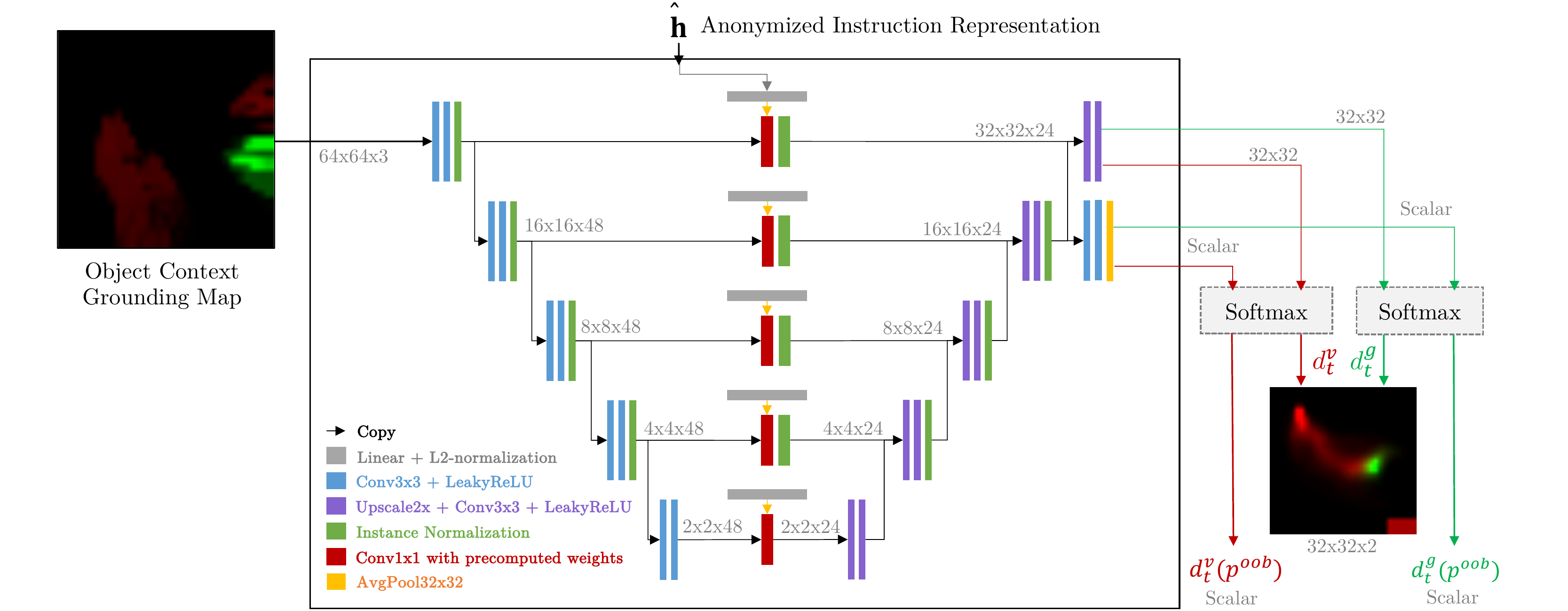}
\caption{The $\lingunet$ architecture. $\lingunet$ outputs raw scores, which we normalize over the domain of each distribution. This figure is adapted from~\citet{blukis2019learning}.}
\label{fig:lingunet}
\vspace{10pt}
\end{figure}

This section is adapted from \citet{blukis2019learning} and is included here for documentation completeness.
Figure~\ref{fig:lingunet} illustrates the $\lingunet$ architecture. 

$\lingunet$ uses a series of convolution and scaling operations.
The input object context map $\contextmapworld_\idxtimestep$ at time $\idxtimestep$ is processed through $L$ cascaded convolutional layers to generate a sequence of feature maps $\featmap_k = \conv^{D}_k(\featmap_{k-1})$, $k = 1\dots L$.
Each  $\featmap_k$ is filtered with a 1$\times$1 convolution with weights $\kernel_k$. The kernels $\kernel_k$ are computed from the object-independent instruction representation $\mathbf{\hat{h}}$ using a learned linear transformation $\kernel_k = \weights^{u}_{k} \mathbf{\hat{h}} + \bias^{u}_{k}$. 
This generates $l$ language-conditioned feature maps $\featmaptxt_k = \featmap_k \circledast \kernel_k$, $k = 1\dots L$.
A series of $L$ upscale and convolution operations computes $L$ feature maps of increasing size:\footnote{$[\cdot,\cdot]$ denotes concatenation along the channel dimension.}

\begin{small}
\begin{eqnarray*}
\featmapdeconv_k = \left\{\begin{array}{lr}
        {\upscale(\conv^U_k([\featmapdeconv_{k+1}, \featmaptxt_{k}]))}, & \text{if } 1 \leq k \leq L-1\\
        {\upscale(\conv^U_k(\featmaptxt_{k}))}, & \text{if } k=L
        \end{array}\right. \;\;.
\end{eqnarray*}
\end{small}

An additional output head is used to output a vector $\lingunetvecout$: 

\begin{small}
\begin{equation}
    \lingunetvecout = \avgpool(\conv^\lingunetvecout(\featmapdeconv_2))\;\;,\nonumber
\end{equation}
\end{small}

where $\avgpool$ takes the average across the spatial dimensions. $\lingunetvecout$ is the logit score assigned to the dummy location $\posoob$ representing all unobserved environment positions.

The output of $\lingunet$ is a tuple ($\featmapdeconv_1$, $\lingunetvecout$), where $\featmapdeconv_1$ is of size $W_w \times H_w \times 2$ and $\lingunetvecout$ is a vector of length 2. We apply a softmax operation across the spatial dimensions to produce the position visitation and goal visitation distributions given ($\featmapdeconv_1$, $\lingunetvecout$).

\subsection{Visitation Distribution Image Encoding}\label{app:model:distribution-rep}

The visitation distributions $\trajvisit_\idxtimestep$ and $\stopvisit_\idxtimestep$ are represented by a four-channel square-shaped tensor of two spatial dimensions over the environment locations.
Two channels correspond to the spatial domain of $\trajvisit_\idxtimestep$ and $\stopvisit_\idxtimestep$, and the other two channels are filled with uniform values $\trajvisit_\idxtimestep(\posoob)$ and $\stopvisit_\idxtimestep(\posoob)$. This four-channel encoding differs from the representation of \citet{blukis2019learning}.

\subsection{Control Network Architecture}\label{app:model:control-network}

The second policy stage $\stageB(\cdot)$  generates the output velocities. It is implemented by a \emph{control network} that receives from the first stage the visitation distribution encoding (Section~\ref{app:model:distribution-rep}), an observability mask $\maskworld_\idxtimestep$, and a boundary mask $\boundaryworld_\idxtimestep$.
The observability mask identifies locations in the environment that have been seen by the agent so far. 
The boundary mask indicates the four environment boundaries.

\begin{figure}
\centering
\includegraphics[scale=0.4]{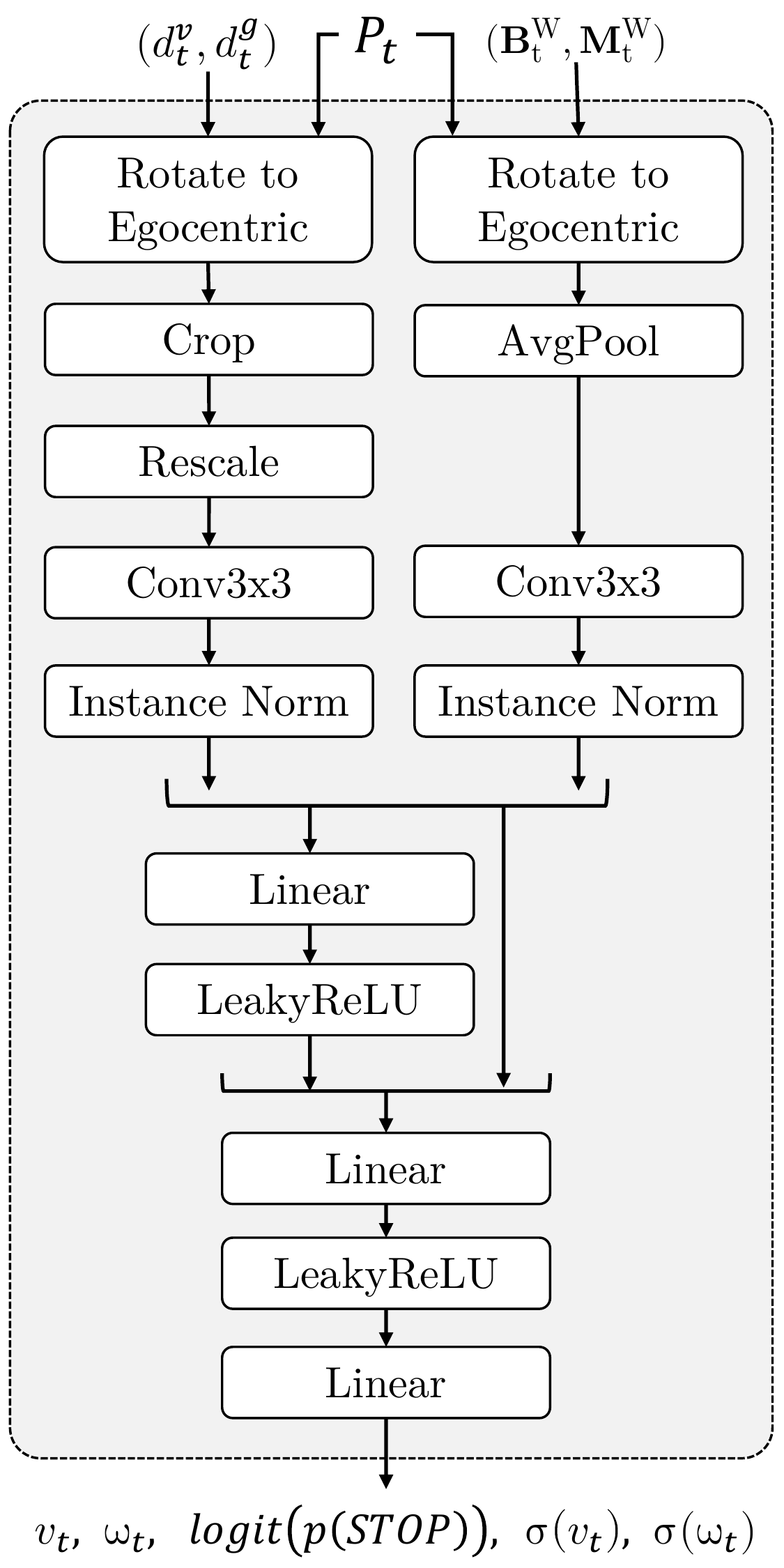}
\caption{Control network architecture.}
\label{fig:control-network}
\end{figure}

Figure~\ref{fig:control-network} shows the control network architecture based on \citet{blukis2019learning}.
We use two distinct copies of the control network, one as the policy Stage 2 action generator, and one as the value function for reinforcement learning as part of the $\rail$ algorithm.

The visitation distributions $\trajvisit_\idxtimestep$ and $\stopvisit_\idxtimestep$ are represented by an image as described in Section~\ref{app:model:distribution-rep}.
This image is rotated to an egocentric reference frame defined by the agent's current pose $\pose_\idxtimestep$, cropped to the agent's immediate area, and processed with a convolutional neural network (CNN).
The observability mask $\maskworld_\idxtimestep$ and boundary mask $\boundaryworld_\idxtimestep$ are concatenated along the channel dimension, spatially resized to the same pixel dimensions as the cropped visitation distributions, and processed with a convolutional neural network (CNN).
The resulting representations of visitation distributions and masks are flattened, concatenated and processed with a densely-connected multi-layer perceptron.

The output of the action generation network consists of five scalars: The predicted forward and angular velocities $\velfwd$ and $\velang$, the logit of the stopping probability, and two standard deviations used during PPO training to define a continuous Gaussian probability distribution over actions.

\section{Learning Details}\label{app:learning}

We train our model with $\sureal$~\cite{blukis2019learning}.
We remove the domain-adversarial discriminator.
The algorithm's two concurrent processes are described  in Algorithms~\ref{algo:supervised} and~\ref{algo:rl}. The pseudocode is adapted from \citet{blukis2019learning} for documentation completeness. 

\paragraph{Process A: Supervised Learning}
Algorithm~\ref{algo:supervised} shows the supervised learning process that is used to estimate the parameters $\paramsA$ of the first policy stage  $\stageA$.
At every iteration, we sample executions from the dataset $\dataset^{\domainsim}$ (line~\ref{ln:sl:sampletraj}) and update using the ADAM~\cite{kingma2014adam} optimizer (line~\ref{ln:sl:modelupdate}) by optimizing the KL-divergence loss function:

\begin{small}
\begin{equation}
\suploss(\execution) = 
     \frac{1}{|\execution|}\sum_{\context \in \gencontexts(\execution)}D_{\rm KL}(\stageA(\context) \| \stageAExpert(\context)) %
\label{eq:suploss}
\end{equation}
\end{small}

\noindent
where $\gencontexts(\execution)$ is the sequence of agent contexts observed during an execution $\execution$,  $\stageAExpert(\context)$ creates the gold-standard visitation distributions (i.e., Stage 1 outputs) for a context $\context$ from the training data, and $D_{\rm KL}$ is the KL-divergence operator.
Every $K^{\rm SL}_{\rm iter}$ iterations, we send the policy stage 1 parameters to Process B (lines~\ref{ln:sl:send}).

\paragraph{Process B: Reinforcement Learning}
Algorithm~\ref{algo:rl} shows the reinforcement learning process that is used to estimate the parameters $\paramsB$ for the second policy stage $\stageB$.
This procedure is identical to the one described in \citet{blukis2019learning} and uses Proximal Policy Optimization~\cite[PPO;][]{schulman2017proximal} to optimize an intrinsic reward function.
At every iteration, we collect $N$ executions by rolling out the full policy $\stageB(\stageAsim(\cdot))$ in the simulator (line~\ref{ln:rl:exec}).
We then perform $K^{\rm RL}_{\rm steps}$ parameter updates optimizing the PPO clipped policy-gradient loss (lines~\ref{ln:rl:ppobegin}-\ref{ln:rl:ppoend}).
We add the collected trajectories to the dataset shared with Process A (line~\ref{ln:rl:mergeexec}).
This allows the first policy stage $\stageA$ to adapt its predicted distributions to the state-visitation distributions induced by the entire policy, making it robust to actions sampled from $\stageB$.
We find that this data sharing prevents $\stageB$ from learning degenerative behaviors that exploit $\stageA$.

We use the same intrinsic reward function as~\citet{blukis2019learning}:

\begin{small}
\begin{equation}
\reward(\context_\idxtimestep, \action_\idxtimestep) =
     \visitweight \visitreward(\context_\idxtimestep, \action_\idxtimestep) + \stopweight \stopreward(\context_\idxtimestep, \action_\idxtimestep) 
     +  \exploreweight \explorereward(\context_\idxtimestep, \action_\idxtimestep) - 
     \actionweight \actionreward(\action_\idxtimestep) - \stepweight\;\;, \label{eq:reward}
\end{equation}
\end{small}

\noindent
where all $\lambda_{(\cdot)}$'s are constant hyperparameter weights, $\visitreward$ rewards correctly following the predicted trajectory distribution, $\stopreward$ rewards stopping at or near a likely stopping position according to the stopping distribution, $\explorereward$ rewards exploring the environment, and $\actionreward$ penalizes actions outside of controller range.
See \citet{blukis2019learning} the formal definition of these terms.

\begin{figure}
\begin{minipage}[t]{\columnwidth}

\begin{algorithm}[H]
\caption{Process A: Supervised Learning}
\begin{algorithmic}[1]
\footnotesize
\Require First stage model $\stageA$ with parameters $\paramsA$, 
dataset of simulated demonstration trajectories ${\dataset^{\domainsim}}$.
\Definitions $\stageA^B$ are shared with Process B.
\State $j \gets 0$
\Repeat
    \State $j \gets j + 1$
    \For{$i = 1, \dots , K^{\rm SL}_{\rm iter}$}
        \State \Comment{Sample trajectory}
        \State $\execution^{\domainsim} \sim \dataset^{\domainsim}$
        \label{ln:sl:sampletraj}
        \State \Comment{Update first stage parameters}
        \State $\paramsA \leftarrow \textsc{ADAM}(\nabla_{\paramsA}\suploss(\execution))$
        \label{ln:sl:modelupdate}
    \EndFor
    \State \Comment{Send $\stageAsim$ to Process B if it is running} 
    \State $\stageAsim^B \gets \stageAsim$ \label{ln:sl:updateA}\label{ln:sl:send}
    \If{$j = K_{\rm iter}^{B}$}
        \State Launch Process B (Algorithm~\ref{algo:rl}) \label{ln:sl:launchB}
        \vspace{-2pt}
    \EndIf
\Until{Process B is finished} \\
\Return $\stageAreal$
\vspace{-3pt}
\end{algorithmic}
\label{algo:supervised}
\end{algorithm}

\begin{algorithm}[H]
\caption{Process B: Reinforcement Learning}
\begin{algorithmic}[1]
\footnotesize
\Require Simulation dataset ${\dataset^{\domainsim}}$, second-stage model $\stageB$ with parameters $\paramsB$, value function $\valuefunc$ with parameters $\valueparams$, first-stage simulation model $\stageAsim$. 
\Definitions $\textsc{Merge}(\dataset, E)$ is a set of sentence-execution pairs including all instructions from $\dataset$, where each instruction is paired with an execution from $E$, or $\dataset$ if not in $E$. $\dataset^\domainsim$ and $\stageAsim^B$ are shared with Process A.

\For{$e = 1,  \dots, K^{\rm RL}_{\rm epoch}$}
    \State \Comment{Get the most recent update from Process A}
    \State $\stageAsim \gets \stageAsim^B $
    \For{$i = 1,  \dots , K^{\rm RL}_{\rm iter}$}
        \State \Comment{Sample simulator executions of $N$ instructions}
        \State $\execposseq^{(1)}, ..., \execposseq^{(N)} \sim \stageB(\stageAsim(\cdot))$ \label{ln:rl:exec}
        \For{$j = 1, \dots, K^{\rm RL}_{\rm steps}$} \label{ln:rl:ppobegin}
            \State \Comment{Sample state-action-return tuples and update} 
            \State $X \sim \execposseq_{1}, ..., \execposseq_{N}$
            \State $\paramsB, \valueparams \leftarrow \textsc{ADAM}(\nabla_{\paramsB,\valueparams}\mathcal{L}_{PPO}(X,  \valuefunc))$
        \EndFor \label{ln:rl:ppoend}
        \State \Comment{Update executions to share with Process A}
        \State $\dataset^{\domainsim} \leftarrow \textsc{Merge}(\dataset^{\domainsim}, \{\execposseq_{1}, \dots , \execposseq_{N}\})$ \label{ln:rl:mergeexec}
    \EndFor
\EndFor\\
\Return $\stageB$
\end{algorithmic}
\label{algo:rl}
\end{algorithm}

\end{minipage}
\end{figure}

\section{Extracting Object References from a Navigation Corpus}\label{app:alignments}

\newcommand{\chunker}{\textsc{Ch}}
\newcommand{\objset}{\textsc{Ob}}
\newcommand{\layout}{\Lambda}
\newcommand{\traj}{\Xi}

We assume access to a dataset $\{(u^{(i)}, \traj^{(i)}, \layout^{(i)})\}_{i}$ of natural language instructions $u^{(i)}$, each paired with a demonstration trajectory $\traj^{(i)}$ and an environment layout $\layout^{(i)}$ that is a set of objects and their poses in the environment.

Given a natural language instruction $u$, let $\chunker(u)$ denote the multi-set of noun chunks that appear in the instruction.
This includes object references, such as \emph{the blue box}, and spurious noun chunks, such as \emph{the left}.
Let $\objset(\layout, \traj)$ denote the set of objects that appear in the layout $\layout$ in the proximity of the trajectory $\traj$, which we define as  within $1.41m$.
We assume that the noun chunks $\chunker(u)$ describe a subset  of objects $\objset(\layout, \traj)$ and use an alignment model similar to IBM Model 1~\cite{brown1993mathematics} to estimate the probabilities $p_{\gamma}(r \mid o)$ for a phrase $r \in \chunker(u)$ and an object $o \in \objset(\layout, \traj)$. The distribution is parameterized by $\gamma$, and is implemented with a one-layer long short-term memory network~\cite[LSTM;][]{hochreiter1997long}. The input is a one-hot vector indicating the object type. The output is a sequence of tokens. The vocabulary is a union of all words in all training instructions.
Noun chunks that do not refer to any landmark (e.g., \nlstring{left side}, \nlstring{full stop}, \nlstring{the front}) are aligned with a special \textsc{Null} object.

Given a noun chunk $r$, we use the alignment model $p_{\gamma}(r \mid o)$  to infer the object $o$ referred by $r$:

\begin{small}
\begin{equation}
    o =  \arg\max_{o \in \objset(\layout, \traj)} p_{\gamma} (r \mid o)p(o)\;\;,
    \label{eq:align:probe}
\end{equation}
\end{small}

\noindent
where $p(o)$ is estimated using object frequency counts in training data.
We use this process to extract a dataset of over 4,000 textual object references, each paired with an object label.
The language includes diverse ways of referring to the same object, such as \nlstring{the barrel}, \nlstring{the lighter colored barrel}, \nlstring{the silver barrel}, \nlstring{white cylinder}, and \nlstring{white drum}.
This technique is applicable to any vision and language navigation dataset that includes object annotations, such as the commonly used R2R dataset~\cite{anderson2017vision}.

\eat{
We train $p_{\gamma}(r \mid o)$ by maximizing Equation~\ref{eq:align:main} on the \textsc{Lani} training data.
We use this process to extract a dataset of over 4,000 textual object references, each paired with an object label.
The language includes diverse ways of referring to the same object, such as \nlstring{the barrel}, \nlstring{the lighter colored barrel}, \nlstring{the silver barrel}, \nlstring{white cylinder}, and \nlstring{white drum}.
This technique is applicable to any vision and language navigation dataset that includes object annotations, such as the commonly used R2R dataset~\cite{anderson2017vision}. 

We implement $p_{\gamma}(r \mid o)$ with a one-layer long short-term memory network~\cite[LSTM;][]{hochreiter1997long}. The input is a one-hot vector indicating object type. The out is a sequence of tokens. The vocabulary is a union of all words in all training instructions.
}

\section{Natural Language Navigation Data Details}\label{app:data:corpora}

We use the natural language instruction data from \citet{misra2018mapping} and \citet{blukis2019learning} for training, and collect additional data with new objects. 
Table~\ref{tab:dataset_sizes} shows basic statistics for all the data available to us. 
Table~\ref{tab:data_splits} summarizes how we used this data in our different experiments. 
The $\fspvn$ and $\pvntwoseen$ models were trained on the ``Train Seen'' data split that includes data with 63 objects. The instructions used to train the policy include the 15 seen objects. This data excludes the eight unseen objects.  The language-conditioned segmentation component is pre-trained on AR data, and is never tuned to adapt to the visual appearance of any of these objects.
The $\pvntwoall$ model was trained on the ``Train All'' data split that includes 71 objects, including the 15 seen and eight unseen objects.
The development results on eight new objects were obtained by evaluating on the ``Dev Unseen'' data split.
The test results on 8 new objects were obtained by evaluating on the ``Test Unseen'' data split.
The test results on 15 previously seen objects were obtained by evaluating on the ``Test Seen'' data split.
We restrict the number of instructions in development and test datasets to a realistic scale for physical quadcopter experiments.

\begin{table}[t]
\begin{center}
\footnotesize
\centering
\begin{tabular}{@{}l l l l l l l l@{}}
\toprule
\multirow{2}{*}{Dataset} & \multirow{2}{*}{Type} & \multirow{2}{*}{Split} & \multirow{2}{*}{\# Paragraphs} & \# Instr. & Avg. Instr. Len. & Avg. Path \\
                         &                       &                        &                                &           & (tokens)   & Len. (m) \\
\cmidrule{1-8}

\multirow{6}{*}{\textsc{Lani}} &
\multirow{3}{*}{(A) 1-segment} &
(a) Train & 4200  &  19762  &  11.04  &  1.53\\
& & (b) Dev & 898  &  4182  &  10.94  &  1.54\\
& & (c) Test &  902  &  4260  &  11.23  &  1.53\\
\cmidrule{2-8}
& \multirow{3}{*}{(B) 2-segment} &
(a) Train & 4200  &  15919  &  21.84  &  3.07\\
& & (b) Dev & 898  &  3366  &  21.65  &  3.10\\
& & (c) Test & 902  &  3432  &  22.26  &  3.07\\
\cmidrule{1-8}
\multirow{6}{*}{\textsc{Real}} &
\multirow{3}{*}{(A) 1-segment} &
(a) Train & 698  &  3245  &  11.10  &  1.00\\
& & (b) Dev & 150  &  640  &  11.47  &  1.06\\
& & (c) Test &  149  &  672  &  11.31  &  1.06\\
\cmidrule{2-8}
& \multirow{3}{*}{(B) 2-segment} &
(a) Train & 698  &  2582  &  20.97  &  1.91\\
& & (b) Dev & 150  &  501  &  21.42  &  2.01\\
& & (c) Test &  149  &  531  &  21.28  &  1.99\\
\cmidrule{1-8}
\multirow{6}{*}{\textsc{Unseen}} &
\multirow{3}{*}{(A) 1-segment} &
(a) Train & 692  &  2790  &  13.60  &  1.20\\
& & (b) Dev & 147  &  622  &  13.41  &  1.16\\
& & (c) Test &  147  &  577  &  13.14  &  1.25\\
\cmidrule{2-8}
& \multirow{3}{*}{(B) 2-segment} &
(a) Train & 692  &  2106  &  25.39  &  2.28\\
& & (b) Dev & 147  &  476  &  24.87  &  2.17\\
& & (c) Test &  147  &  431  &  24.77  &  2.39\\
\bottomrule
\end{tabular}
\caption{Dataset and split sizes. \textsc{Lani} was introduced by \citet{misra2018mapping} and contains a total of 63 different objects in simulation only. \textsc{Real} is additional data introduced by~\citet{blukis2019learning} with 15 objects that are a subset of \textsc{Lani} objects for use on the physical drone or simulation.
\textsc{Unseen} is data that we collected containing environments with only 8 new objects that did not appear in \textsc{Lani} or \textsc{Real} data. It allows us to train models on data from \textsc{Lani} and \textsc{Real}, while testing on data with previously unseen objects from \textsc{Unseen}. The 2-segment data consists of instructions made of two 1-segment consecutive instructions concatenated together.}
\label{tab:dataset_sizes}
\end{center}
\end{table}

\begin{table}[t]
\begin{center}
\footnotesize
\centering
\begin{tabular}{@{}l l l@{}}
\toprule
Data Split & \# Instr. & Source from data splits in Table~\ref{tab:dataset_sizes}.\\
\cmidrule{1-3}
Train Seen & 41508 & \textsc{Lani}.A.a $\cup$ \textsc{Lani}.B.a $\cup$ \textsc{Real}.A.a $\cup$ \textsc{Real}.B.a \\
Train All & 46404 & \textsc{Lani}.A.a $\cup$ \textsc{Lani}.B.a $\cup$ \textsc{Real}.A.a $\cup$ \textsc{Real}.B.a $\cup$ \textsc{Unseen}.A.a $\cup$ \textsc{Unseen}.B.a\\
\cmidrule{1-3}
Dev Unseen &  103  &  Random subset of \textsc{Unseen}.B.b \\
Test Unseen &  63  &  Random subset of \textsc{Unseen}.B.c \\
Test Seen &  73  &  Random subset of \textsc{Real}.B.c \\
\bottomrule
\end{tabular}
\caption{Dataset splits used for training, development and testing in our experiments in Section~\ref{sec:results}, showing the number of instructions, and how each data split was obtained from the available data summarized in Table~\ref{tab:dataset_sizes}}
\label{tab:data_splits}
\end{center}
\end{table}

\subsection{List of Seen and Unseen Objects}\label{app:data:corpora:objects}

\begin{figure}[t]
\centering
\includegraphics[scale=0.4]{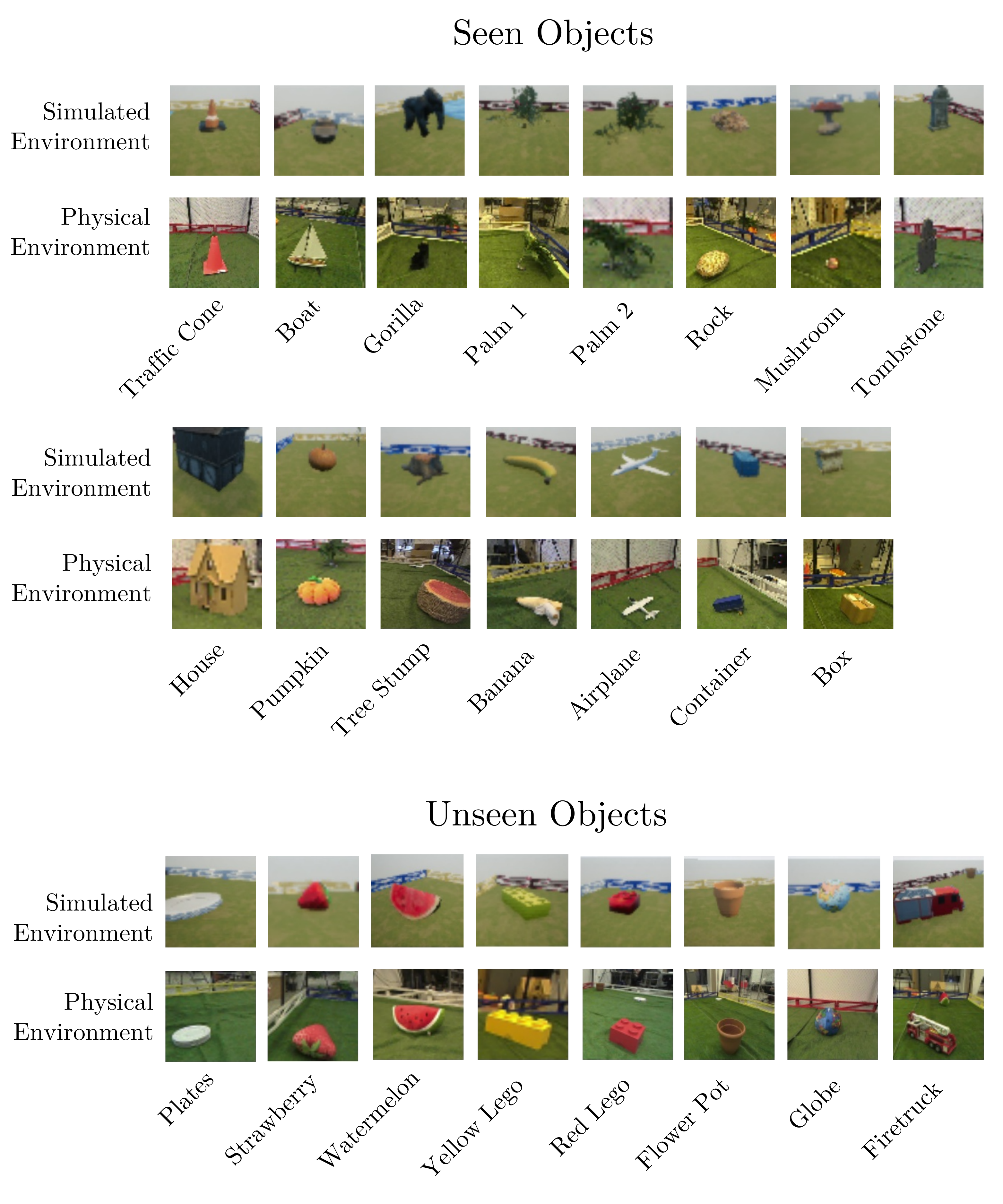}
\caption{The list of seen (top) and unseen (bottom) objects  during training in both the physical and real-world environments.}
\label{app:fig:object_lists}
\end{figure}

Figure~\ref{app:fig:object_lists} shows the set of seen and unseen objects in the simulated and physical environments. An additional 48 simulation-only objects that are seen during training are not shown.
The agent does not see the unseen objects or references to them during training. 

\section{Augmented Reality Object Image Data}
\label{app:ardata}

\begin{figure}
\centering
\includegraphics[scale=0.4]{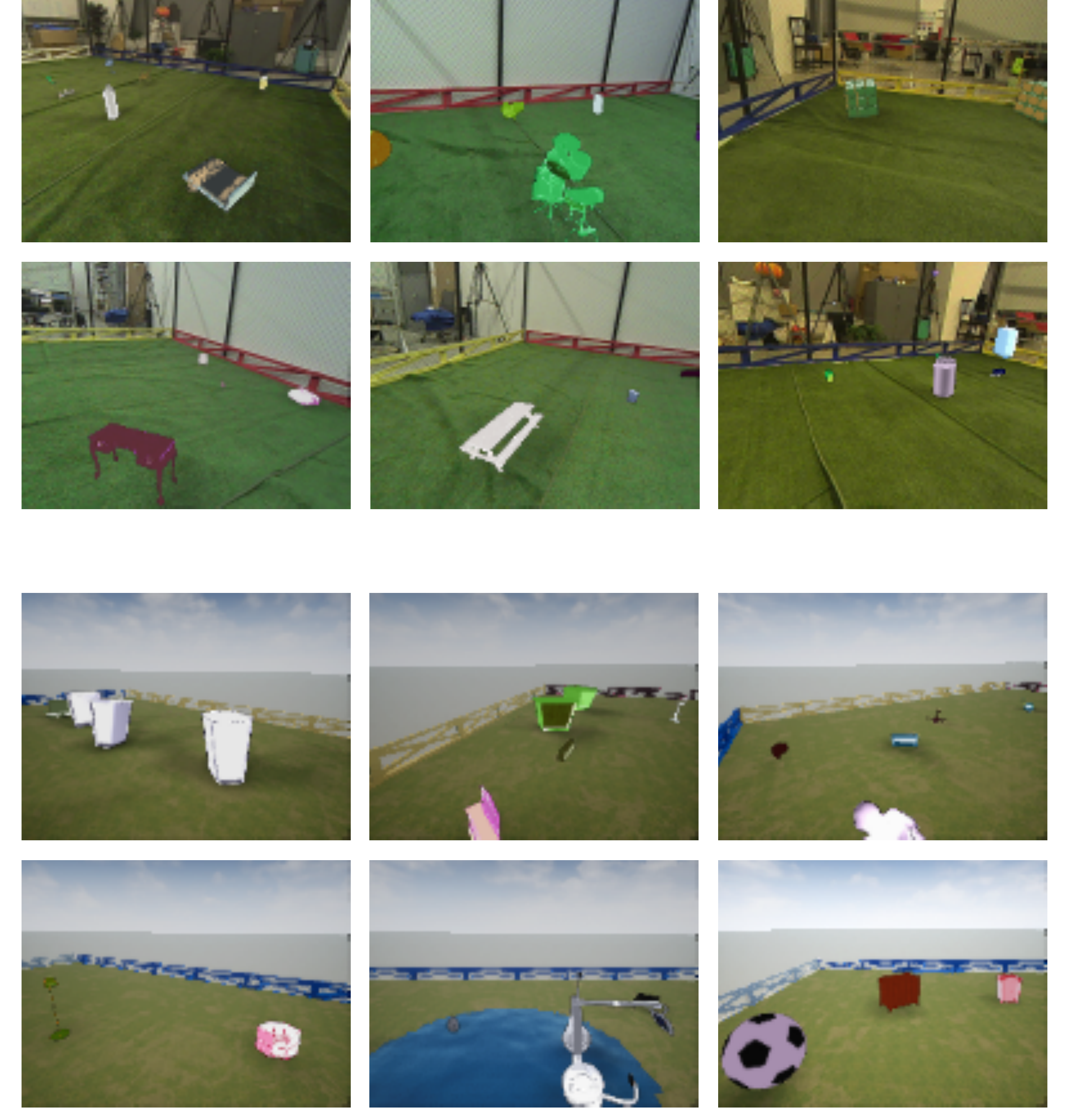}
\caption{Examples from the augmented reality object data in the physical (top) and simulated (bottom) environments.}
\label{fig:app:data:ardata}
\end{figure}

The training procedure for the few-shot language-conditioned segmentation component uses a dataset $\objectdata = \{(\image^{(i)},\{ (\bbox^{(i)}_j, \objmask^{(i)}_j, \object^{(i)}_j) \}_j)\}_{i}$  (Section~\ref{sec:segment:learning}).
This data includes a large number of diverse objects that cover general object appearance properties, such as shape, colors, textures, and size, to allow generalizing to new objects.
Collecting such data in a real-world environment is costly.
Instead, we generate 20,000 environment layouts that consist of  6--16 objects drawn from a pool of 7,441 3D models from ShapeNet~\cite{chang2015shapenet}. We use only objects where the longest edge of the axis-aligned bounding box is less than five times grater than the shorter edge. This excludes planar objects such as paintings. 
We use augmented reality to instantiate the objects in the physical and simulated environments.
We collect a set of images of empty environments with no objects by flying the quadcopter along randomly generated trajectories.
We  use the Panda3D~\cite{goslin2004panda3d} rendering engine to render objects over the observed images, as if they are physically present in the environment.
Figure~\ref{fig:app:data:ardata} shows example observations.
This process also automatically generates bounding box annotations tagged with object identities.
The diverse shapes and textures of objects allow us to learn a view-point invariant object similarity metric, however it creates the challenge of generalizing from relatively simple ShapeNet objects to physical objects.
It is possible to load the ShapeNet objects within the simulator itself, but we opted to use the AR approach in both simulated and physical environments to ensure uniform data format.

\eat{
The construct an augmented reality dataset $\objectdata = \{(\image^{(i)},\{ (\bbox^{(i)}_j, \objmask^{(i)}_j, \object^{(i)}_j \}_j)\}_{i}$, where $\image^{(i)}$ is a first-person image and $\bbox^{(i)}_j$ is a bounding box of the object $\object^{(i)}_j$, which has the object mask $\objmask^{(i)}_j$. Figure~\ref{fig:app:data:ardata} shows examples of images, bounding boxes, and masks.

We initialize an empty environment with no objects, and fly the quadcopter with a non-learning policy to collect a set of random trajectories $\langle (\image_1, \pose_1), \dots, (\image_{\execlen}, \pose_{\execlen}) \rangle$, where $\image_i$ is the first-person image observed by the agent and $\pose_i$ is the corresponding camera pose.
We used the $\textsc{Oracle}$ policy to follow randomly drawn human demonstrations from the training data.
The output of this process is a set $\{\image_j, \pose_j\}_{j=1}^{}$ of first-person images from the drone flight perspective in empty environments.

Next, we use the Panda3D\footnote{http://www.panda3d.org/} rendering engine to instantiate random layouts of ShapeNet\cite{chang2015shapenet} objects within the environment bounds.
We filter out 
For each image $\image_j$ with pose $\pose_j$ render an AR overlay image $\image^\Delta_j$, such that $\image_j + \image^\Delta_j$ produces 

This data can be generated in the simulated and real environment using the same process.
Theoretically the simulation environment

We generate this data by placing 3D objects in the simulated environment or overlaying them on images from the physical environment. 
}

\section{Object Databases}\label{app:objectdata}

\begin{figure*}[t]
\centering
\includegraphics[scale=0.4]{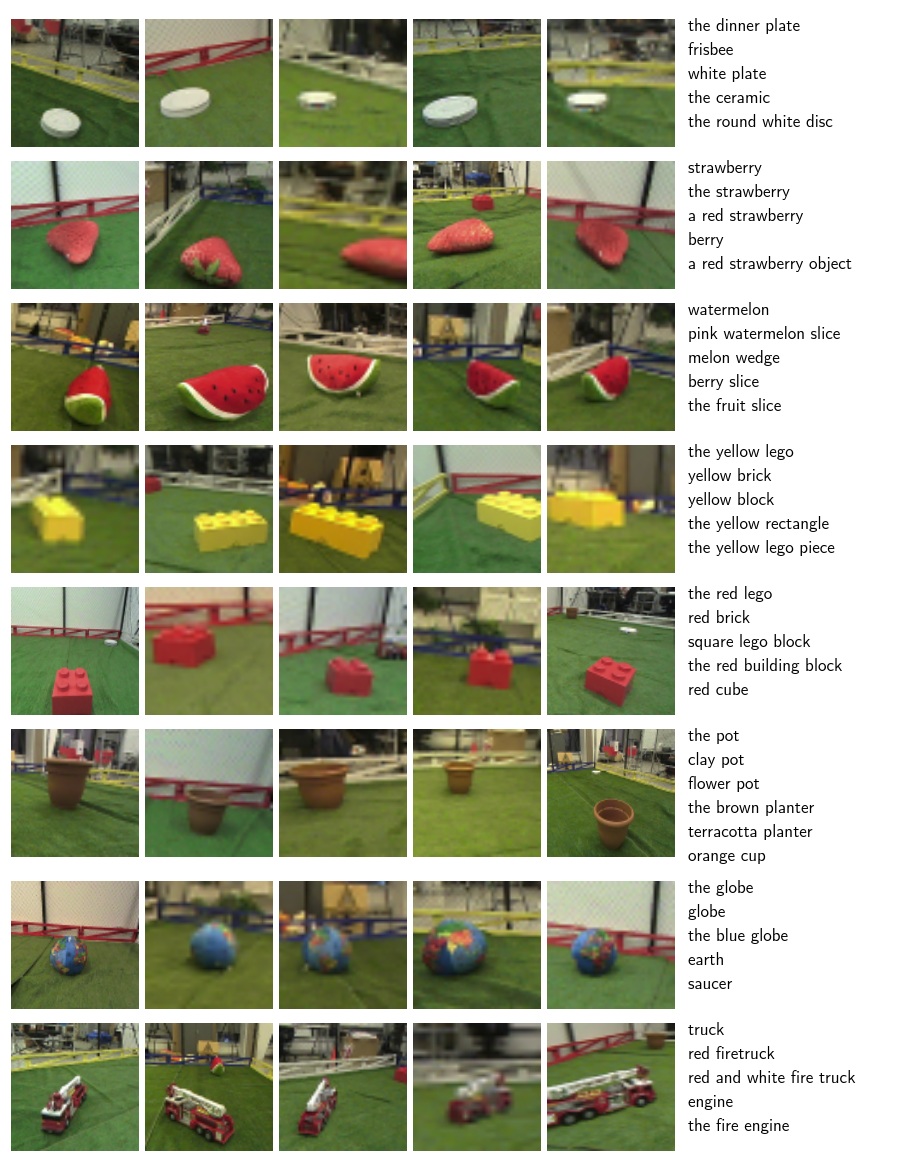}
\caption{The object database used during testing, containing previously unseen physical objects.}
\label{fig:objectdata_real_unseen}
\end{figure*}

\begin{figure*}[t]
\centering
\includegraphics[scale=0.4]{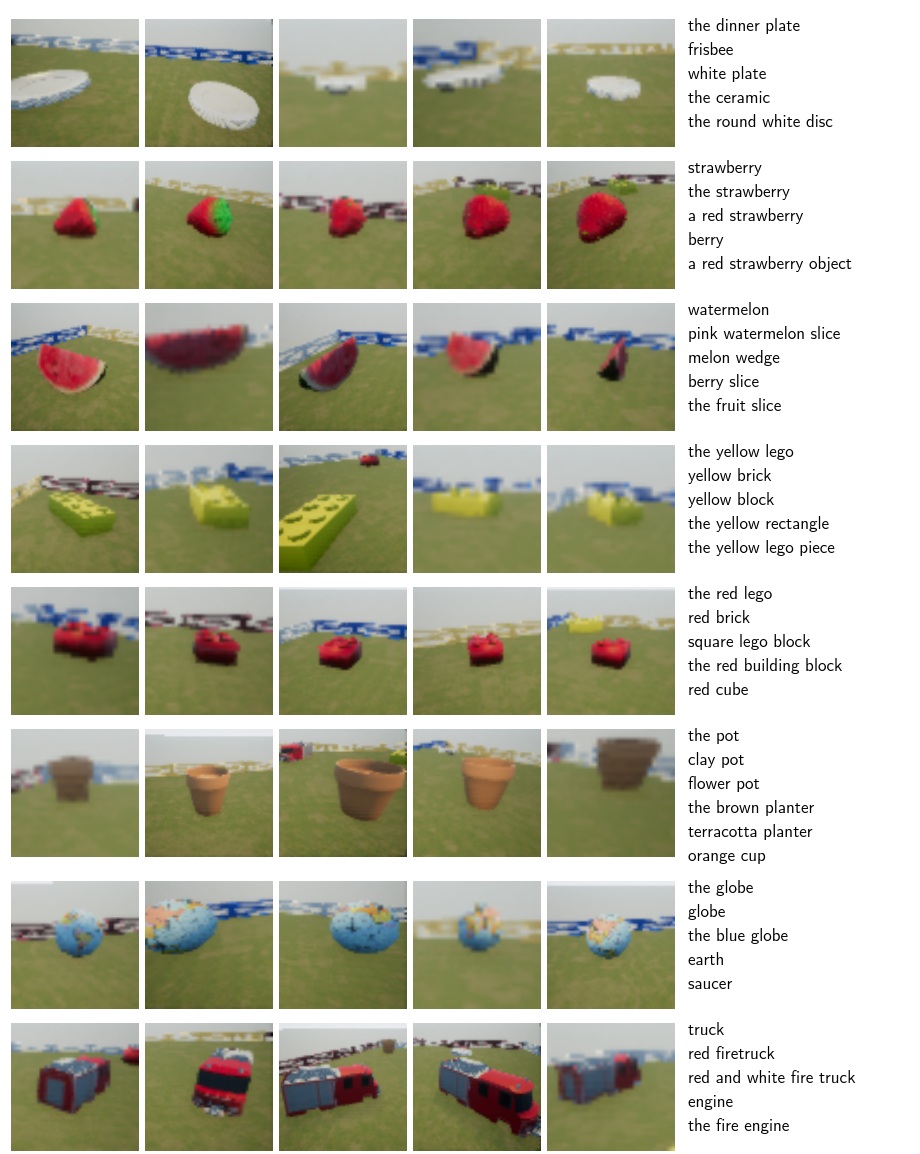}
\caption{The object database  used during testing, containing previously unseen simulated objects.}
\label{fig:objectdata_sim_unseen}
\end{figure*}

\begin{figure*}[t]
\centering
\includegraphics[scale=0.3]{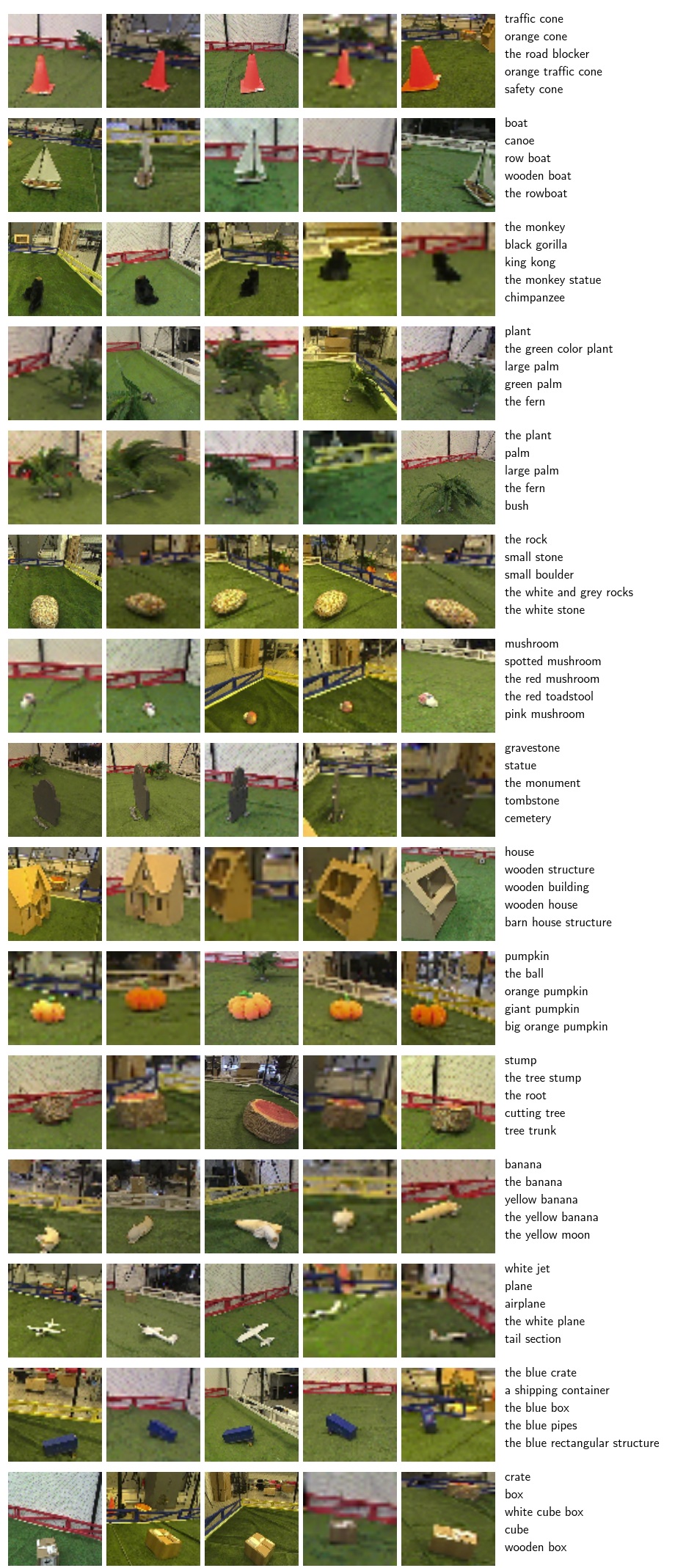}
\caption{The object database used during development in the physical environment.}
\label{fig:objectdata_real_seen}
\end{figure*}

\begin{figure*}[t]
\centering
\includegraphics[scale=0.3]{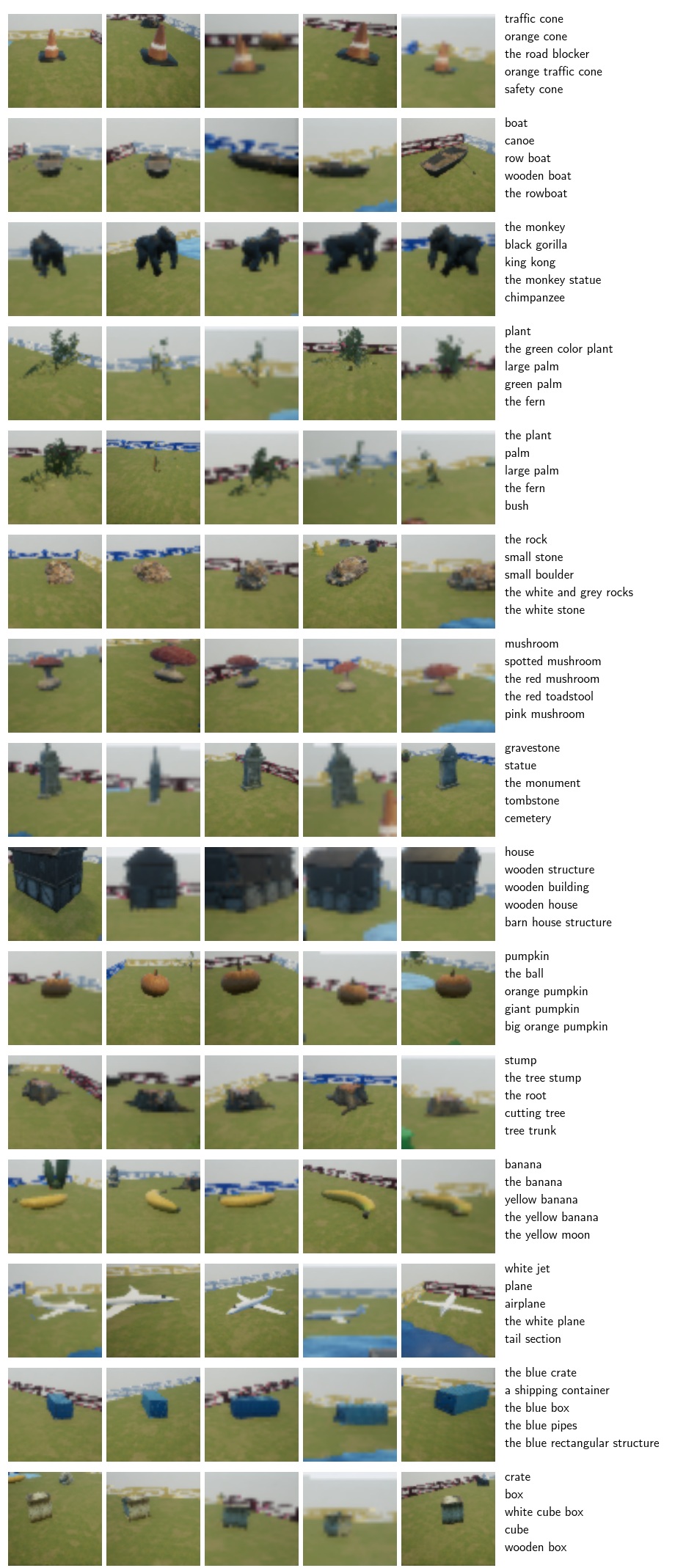}
\caption{The object database  used during development in the simulation..}
\label{fig:objectdata_sim_seen}
\end{figure*}

The object database $\nod$ consists of a set of objects, each represented by five images and five textual descriptions.
We use different object databases during training, development evaluation on seen objects, and test-time evaluation on unseen object.
For each database, the images and textual descriptions are taken from a pre-collected pool.
Object images are obtained by collecting a set of trajectories from the $\oracle$ policy in random training environments, cropping out a region around each object in the image, and storing each image tagged by the object type.
The textual description are obtained by first extracting all noun chunks in every instruction of the $\lani$ dataset training split using SpaCy~\cite{spacy2}, and using Equation~\ref{eq:align:probe} to match each noun chunk to the object it refers to.

\subsection{Object Database Figures}

\paragraph{Test-time Database for Evaluation with Unseen Objects}

Figure~\ref{fig:objectdata_real_unseen} shows the object database used at test-time on the physical quadcopter containing unseen objects only. The agent has not seen these objects before, and the only information it has available about these objects is the database.
The images and textual descriptions are hand-selected from the pre-collected pool to be diverse and representative.
Figure~\ref{fig:objectdata_sim_unseen} shows the same content for the simulation.

\paragraph{Development Database for Evaluation with Seen Objects}

Figure~\ref{fig:objectdata_real_seen} shows the object database used for evaluation during development on the physical quadcopter containing objects that the agent has seen during training.
The images and textual descriptions are hand-selected from the pre-collected pool to be diverse and representative.
Figure~\ref{fig:objectdata_sim_seen} depicts the same content for the simulation.

\paragraph{Generation of Object Databases Used During Training}

Each training example is a tuple $(\instruction,  \execution)$ situated in an environment layout $\Lambda$ that specifies the set of objects in the environment and their poses.
We generate an object database $\nod$  for each training example by creating an entry in the database for each object $\object \in \Lambda$.
We pair each object with five images randomly selected from the pre-collected image pool, and  five object references randomly selected from the pre-collected textual description pool.

\section{Additional Evaluation}
\label{app:results}

\paragraph{Language-Conditioned Segmentation Evaluation}
Automatic evaluation of our language-conditioned segmentation is not possible due to a lack of ground-truth alignments between object references in the instructions and object masks.
We manually evaluate our language-conditioned segmentation method on 40 policy rollouts from the development data containing unseen objects to assess its performance in isolation.
For each rollout, we subjectively score the segmentation mask output with a score of 1--3, where 1 means the output is wrong or missing, 2 means that at least one of the mentioned objects has been identified, and 3 means that all mentioned objects have been correctly identified, allowing only for slight visual artifacts in mask boundaries.
Because each rollout consists of a sequence of images, we allow for some images to contain false negatives, so long as the mentioned objects are eventually identified in a way that conceivably allows the policy to complete the task.
Our approach achieved a 3-point score on 82.5\% of the rollouts.

\paragraph{Image Similarity Measure Evaluation}
We automatically evaluate the image similarity model $\imageembedding$ in isolation on a 2-way, 8-way, and 15-way classification task using 2429 images of 15 physical objects in the drone environment.
We use the set of ``seen'' objects (Figure~\ref{fig:objectdata_real_seen}).
In each evaluation example, we randomly sample a query object with five random query images, and a set of target objects with five random images each. The set of target objects includes the query object, but with a different set of images.
We test the ability of the image similarity model $\imageembedding$ to classify which of the target objects has the same identity as the query object.

We find that in the 2-way classification task (n=11480), the image similarity model achieves 92\% accuracy in identifying the correct target object.
In a 8-way classification task (n=14848) the accuracy drops to 73\%, and on a 15-way classification task (n=14848), it drops to 63\%.
The model has never observed these objects, and generalizes to them from AR training data only.

The language-conditioned few-shot segmentation model combines both visual and language modalities to identify an object and produce an instance segmentation mask, considering every object in the database. This is why the segmentation model that uses $\imageembedding$ can achieve a higher segmentation performance than $\imageembedding$ achieves on a classification task in isolation.

\section{Implementation Details}\label{app:impl}

\subsection{Hyperparameter Settings}
Table~\ref{tab:hyper} shows the hyperparameter assignments.
We started with the initial values from ~\citet{blukis2019learning}, and tuned the parameters relating to our few-shot grounding approach.

\begin{table*}[t]
  \footnotesize
  \centering
  \begin{tabular}{@{} l l @{}}
  \toprule
  \textbf{Hyperparameter} & \textbf{Value} \\
  \cmidrule{1-2}
  \multicolumn{2}{c}{Environment Settings}\\
  \cmidrule{1-2}
Maximum yaw rate & $\velang_{\rm max} = 1m/s$\\
Maximum forward velocity & $\velfwd_{\rm max} = 0.7m/s$\\
  \cmidrule{1-2}
  \multicolumn{2}{c}{Image and Feature Dimensions}\\
  \cmidrule{1-2}
Camera horizontal FOV & $84^{\circ}$\\
Input image dimensions & $128 \times 72 \times3$\\
Object mask $\maskworld$ dimensions & $32 \times 32 \times 1$\\
Object context map $\contextmapworld$ dimensions & $32 \times 32 \times 40$\\
Visitation distributions $\stopvisit$ and $\trajvisit$ dimensions & $64 \times 64 \times 1$\\
Database object image $\query$ dimensions & $32 \times 32 \times 3$\\
Environment edge length in meters & $4.7m$\\
  \cmidrule{1-2}
  \multicolumn{2}{c}{Few-shot Language-Conditioned Segmentation}\\
  \cmidrule{1-2}
Image metric learning margin & $\tripletmargin = 1.0$\\
Image metric learning margin & $\tripletmarginB = 2.0$\\
Image kernel density estimation std. dev. & $\sigma = 2.0$\\
Text kernel density estimation std. dev. & $\sigma = 0.5$\\

Object reference recognizer weight & $\lambda_{R1} = 0.5$\\
Object reference recognizer threshold & $\lambda_{R2} = 0.03$\\

  \cmidrule{1-2}
  \multicolumn{2}{c}{General Learning}\\
  \cmidrule{1-2}
Deep Learning library & PyTorch 1.4.1\\
  \cmidrule{1-2}
  \multicolumn{2}{c}{Supervised Learning}\\
  \cmidrule{1-2}
Optimizer & ADAM \\
Learning Rate & $0.001$\\
Weight Decay & $10^{-6}$\\
Batch Size & $1$ \\
  \cmidrule{1-2}
  \multicolumn{2}{c}{Reinforcement Learning (PPO)}\\
  \cmidrule{1-2}
Num supervised epochs before starting RL ($K_{iter}^{B}$) & 30\\
Num epochs ($K_{\rm epoch}^{\rm RL}$) & 200\\
Iterations per epoch ($K_{\rm iter}^{\rm RL})$ & 50\\
Number of parallel actors & 4\\
Number of rollouts per iteration $N$ & 20\\
PPO clipping parameter & 0.1\\
PPO gradient updates per iter ($K_{\rm steps}^{\rm RL})$ & 8\\
Minibatch size & 2\\
Value loss weight & 1.0\\
Learning rate & 0.00025\\
Epsilon & 1e-5\\
Max gradient norm & 1.0\\
Use generalized advantage estimation & False\\
Discount factor ($\gamma$) & 0.99\\
Entropy coefficient & 0.001\\
  \cmidrule{1-2}
  \multicolumn{2}{c}{Reward Weights}\\
  \cmidrule{1-2}
  Stop reward weight ($\stopweight$) & 0.5\\
    Visitation reward weight($\visitweight$) & 0.3\\
    Exploration reward weight ($\exploreweight$) & 1.0\\
    Negative per-step reward ($\stepweight$) & -0.04\\
  \bottomrule
  \end{tabular}
  \caption{Hyperparameter values.}
  \label{tab:hyper}
\end{table*}

\end{document}